\documentclass[11pt,draftcls,onecolumn]{IEEEtran}
\usepackage{graphicx,subfigure}
\usepackage{epsfig}
\usepackage{cite}
\usepackage{amsmath,amsfonts,amssymb,amsthm}
\usepackage{algorithm, algorithmic}
\usepackage{tikz}
\usepackage{tabularx}
\usepackage{caption}

\theoremstyle{plain}
\newtheorem{thm}{Theorem}
\newtheorem*{thm*}{Theorem}

\newtheorem{lem}{Lemma}
\newtheorem*{lem*}{Lemma}
\newtheorem{cor}{Corollary}
\newtheorem{prop}{Proposition}
\newtheorem*{prop*}{Proposition}

\theoremstyle{definition}

\newtheorem{Def}{Definition}

\newcommand*\circled[1]{\tikz[baseline=(char.base)]{
            \node[shape=circle,draw,inner sep=2pt] (char) {#1};}}

\newcommand{\BigO}[1]{\ensuremath{\operatorname{O}\left(#1\right)}}

\begin{document}
\title{Confidence-constrained joint sparsity  recovery under the Poisson noise model}
\author{E. Chunikhina, R. Raich, and T. Nguyen\footnote{
Department of EECS,  Oregon State University, Corvallis, OR  97331, USA
\texttt{chunikhe,raich,thinhq@eecs.oregonstate.edu}}}
\date{ }
\maketitle

\begin{abstract}
Our work is focused on the joint sparsity recovery problem where the common sparsity pattern is corrupted by Poisson noise. We formulate the confidence-constrained optimization problem in both least squares (LS) and maximum likelihood (ML) frameworks and study the conditions for perfect reconstruction of the original row sparsity and row sparsity pattern. However, the confidence-constrained optimization problem is non-convex. Using convex relaxation, an alternative convex reformulation of the problem is proposed. We evaluate the performance of the proposed approach using simulation results on synthetic data and show the effectiveness of proposed row sparsity and row sparsity pattern recovery framework.
\end{abstract}


\section{Introduction}
\label{sec:Introduction}

\subsection{Background}
\label{ssec:background}

The problem of recovering jointly sparse solutions for inverse problems is receiving increased attention. This problem also known as the multiple measurement vector (MMV) problem is an extension of the single measurement vector (SMV) problem - one of the most fundamental problems in compressed sensing \cite{candes2006compressive}, \cite{donoho2006compressed}. The MMV problem arises naturally in many applications, such as equalization of sparse communication channels \cite{fevrier1999reduced}, \cite{cotter2002sparse}, neuromagnetic imaging \cite{cotter2005sparse}, magnetic resonance imaging with multiple coils \cite{pruessmann1999sense},  source localization \cite{malioutov2005sparse}, distributed compressive sensing \cite{baron2005distributed}, cognitive radio networks \cite{meng2011collaborative}, direction-of-arrival estimation in radar \cite{krim1996two}, feature selection \cite{nie2010efficient}, and many others.

Several methods have been developed to solve the MMV problem. Most notable among them are the forward sequential search-based method \cite{fevrier1999reduced}, ReMBo, that reduces the MMV problem to a series of SVM problems \cite{mishali2008reduce}, CoMBo, that concatenates MMV to a larger dimensional SMV problem \cite{lee2009concatenate}, alternating direction principle method \cite{deng2011group}, methods based on greedy algorithm \cite{cotter2005sparse}, \cite{tropp2006algorithms1}, methods that use convex optimization approach \cite{tropp2006algorithms2},\cite{chen2006theoretical}, methods that use thresholded Landweber algorithm \cite{fornasier2008recovery, teschke2007multi, ramlau2005tikhonov, daubechies2004iterative}, methods that use restricted isometry property (RIP) \cite{candes2006robust, candes2006stable, candes2008restricted}, extensions to FOCUSS class of algorithms \cite{rao1998sparse}. A good review of different recovery methods and uniqueness conditions for MMV were discussed in \cite{lee2009concatenate, rao1998basis, tropp2006algorithms1, tropp2006algorithms2, mishali2008reduce, cotter2005sparse, chen2006theoretical, eldar2008robust, baron2005distributed}.

Although many works have mainly dealt with noiseless case, there are extensions of the MMV problem with the assumption of noise, notably, additive Gaussian noise \cite{cotter2005sparse, mishali2008efficient, silva2006selecting, zhang2010sparse}. However, in many applications such as Magnetic Resonance Imaging (MRI), emission tomography, astronomy and microscopy, the statistics of the noise corrupting the data is described by the Poisson process. The SMV problem with Poisson noise was considered in \cite{raginsky2010compressed, willett2009performance, motamed2013sparse}. To the best of our knowledge, there is no work considering MMV problem with Poisson noise.

We note that the MMV problem with Poisson noise is defined in a statistical setting, hence one may consider maximum likelihood (ML) estimation. However, the classical ML method is more likely to oversmooth solution in the regions where signal changes sparsity. Moreover, ML tends to produce a solution that has a good fit to the observation data, which leads to an incorrectly predicted sparsity. Therefore, it is beneficial to balance classical ML approach with a function that enforces the desired property of the solution.

One of the recently proposed approaches \cite{raginsky2010compressed, raginsky2011performance} is based on the fact that maximizing the expected value of the log-likelihood of the Poisson data is equivalent to minimizing Kullback-Leibler divergence between measured value of the data and the estimated value of the data. Therefore, solving optimization problems whose objective is the sum of the penalized Kullback-Leibler (KL) divergence (or penalized I-divergence) and a penalty function that enforces sparsity allow one to obtain a sparse solution with a good fit to Poisson data.

There are some difficulties with MMV problems under the Poisson noise models, namely: Poisson noise is non-additive, and the variance of the noise is proportional to the intensity of the signal. When solving these optimization problems without explicit consideration of sparsity, the resulted solutions tend to overly smooth across different regions of signal intensities. Moreover, unlike the problems with additive Gaussian noise, Poisson noise models impose additional non-negative constraints on the solutions.

\subsection{Our contributions}
\label{ssec:contributions}

In this paper: 1) we consider joint sparsity recovery problem under Poisson noise model with one interesting extension. We use different mixing matrices to obtain different measurement vectors of Poisson counts. This makes problem of joint sparse recovery even more challenging; 2) we propose the confidence-constrained approach to the optimization formulation via two frameworks, Least Squares (LS) and Maximum Likelihood (ML). The  confidence-constrained approach allows for a tuning-free optimization formulation for both LS and ML frameworks; 3) under the assumption that the mixing matrices satisfy RIP and that noise follows a Poisson distribution, we investigate conditions for perfect reconstruction of the original row sparsity and row sparsity pattern for both frameworks. Specifically, we derive confidence intervals that depend only on the dimension of the problem, corresponding probability of error, and observation, but not on the parameters of the model; and 4) we use convex relaxation to reformulate the original non-convex optimization problem as a tuning free convex optimization.

\subsection{Notations}
\label{ssec:NotationConvention}

\begin{itemize}
  \item We define matrices by uppercase letters and vectors by lowercase letters. All vectors are column vectors. Transpose of a vector $x\in\mathbb{R}^{D}$ is denoted as $x^{\mathrm{T}}.$ For a matrix $X \in \mathbb{R}^{D_1 \times D_2}$, $x_{i}$ denotes its $D_1$-dimensional $i$-th column and $(x^{j})^{\mathrm{T}}$ denotes its $D_2$-dimensional $j$-th row. The $j$-th element of the $i$-th column of matrix $X$ is denoted by $x_{i}(j)$ or $X(j,i).$
  \item A vector $e_i$ is the canonical vector satisfying $e_i(j)=1$ for $j=i$ and $0$ otherwise.
  \item The $\ell_p$ norm of a vector $x\in\mathbb{R}^{D}$, $p\ge 1$ is given as $\|x\|_p = \left(\sum_{i=1}^{D} |x(i)|^p\right)^{1/p}.$
  \item The Frobenius norm of a matrix $X \in \mathbb{R}^{D_1\times D_2}$ is defined as $\|X\|_F = \sqrt{\sum_{j=1}^{D1}\sum_{i=1}^{D2}|X(j,i)|^2}.$
  \item Notation $X\ge 0$ is used to denote a matrix $X \in \mathbb{R}^{D_1\times D_2}$ whose elements are nonnegative, i.e., $X(j,i)\ge 0,$ $\forall j = \overline{1,D_1}$ and $\forall i = \overline{1,D_2}$.
  \item A matrix $X \in \mathbb{R}^{D_1\times D_2}$ satisfies the s-restricted isometry property (RIP), if there exists a constant $s\ge0$ and $\delta_s \in [0,1)$ such that for every $s$-sparse vector $y \in \mathbb{R}^{D_2 \times 1}$,
      \begin{equation}
      \label{rip}
      (1-\delta_s)\|y\|_2 \le \|Xy\|_2 \le (1+\delta_s)\|y\|_2.
      \end{equation}
      The constant $\delta_s$ is called the s-restricted isometry constant.
  \item Function $D_{KL}(\cdot||\cdot):\mathbb{R}_{+}^{D} \times \mathbb{R}_{+}^{D} \to \mathbb{R}_{+}$ is called the Kullback-Leibler divergence and is defined only for vectors $x,y \in \mathbb{R}_{+}^{D}$ such that $\|x\|_1 = 1$, $\|y\|_1 = 1$ as follows:
\begin{equation}
\label{Dkldivdef}
D_{KL}\left(x||y\right) = \left\{
  \begin{array}{ll}
    \sum_{i=1}^{D}x(i)\log\Big(\frac{x(i)}{y(i)}\Big), & \hbox{supp($x$) $\subseteq$ supp($y$);} \\
    + \infty, & \hbox{supp($x$) $\not\subseteq$ supp($y$),}
  \end{array}
\right.
\end{equation}
where $\mathrm{supp}(x) = \{i \in [1,D]:x(i) \ne 0\}.$ For formula (\ref{Dkldivdef}) we use the following assumptions: 1) $\log(0) = -\infty;$ 2) $\log(\frac{a}{0}) = +\infty, a \ne 0;$ 3) $0 \times (\pm \infty) = 0.$
  \item Function $I\left(\cdot||\cdot\right):\mathbb{R}_{+}^{D} \times \mathbb{R}_{+}^{D} \to \mathbb{R}_{+}$ is called the I-divergence (also known as the generalized Kullback-Leibler divergence or Csiszar's divergence) and is defined for vectors $x,y \in \mathbb{R}_{+}^{D}$ as follows:
\begin{equation}
\label{Idivdef}
I\left(x||y\right) = \left\{
  \begin{array}{ll}
    \sum_{i=1}^{D}\Big(x(i)\log\Big(\frac{x(i)}{y(i)}\Big)+y(i)-x(i)\Big), & \hbox{supp($x$) $\subseteq$ supp($y$);} \\
    + \infty, & \hbox{supp($x$) $\not\subseteq$ supp($y$).}
  \end{array}
\right.
\end{equation}
For formula (\ref{Idivdef}) we use the same assumptions as for formula (\ref{Dkldivdef}).
  \item The $\ell_0$ norm of a vector $x \in \mathbb{R}^{D}$ is defined as $\|x\|_0 = \mathrm{card}\{j \in [1,D]: x(j) \ne 0\}$ that is the total number of non-zero elements in a vector. A vector $x$ is $s$-sparse if and only if $\|x\|_0=s$.
  \item The row-support (or row sparsity pattern) of a matrix $X \in \mathbb{R}^{D_1 \times D_2}$ is defined as $\mathrm{RSupp}(X) = \{j \in [1,D_1]: \|(x^{j})^{\mathrm{T}}\|_2 \ne 0]\}.$
  \item For a given matrix $X \in \mathbb{R}^{D_1\times D_2}$ we define $\| X\|_{R^{0}}$ to be the number of rows of matrix $X$ that have non-zero elements, i.e., $\| X\|_{R^{0}} = \mathrm{card} \{\mathrm{RSupp}(X)\}$. A matrix $X$ is $s$ row sparse if $\|X\|_{R^{0}}=s.$ 
  \item Symbol $\mathbb{N}_{0}$ denotes the set of all natural numbers with zero.
\end{itemize}

\section{Problem formulation: the row sparsity recovery under the Poisson noise model}
\label{sec:ProblemFormulation}

Let $\tilde X \in \mathbb{R}_{+}^{K\times N}$ be the true, $s$-row sparse matrix, i.e., each column vector $\tilde x_{i}$ is $s$-sparse and all column vectors have a common sparsity pattern, meaning that indices of nonzero elements of $\tilde x_{i}$ are the same $\forall i=\overline{1,N}$.
The mixing matrices $A_i\in\mathbb{R}_{+}^{M\times K}$ are known and are different for each measurement, i.e., $A_i \ne A_j$, $i\ne j$. The measurement matrix of observed Poisson counts $Y \in \mathbb{N}_0^{M\times N}$ is known and is obtained as follows:
\begin{equation}
\label{Poisson}
y_i(j) \sim \mathrm{Poisson}(\lambda_{i}(j)), \forall i=\overline{1,N}, \forall j=\overline{1,M},
\end{equation}
where $\lambda_{i}=A_i \tilde x_i, \forall i=\overline{1,N}.$

We denote $N\geqslant1$ to be the number of measurements. Note that when $N=1$ problem (\ref{Poisson}) becomes a single measurement vector (SMV) problem. We assume $N\ll M \ll K$. The last assumption indicates that we have a set of under-determined equations, because the number of columns of the $A_i$ matrices is greater than the number of rows. In general, this setting may lead to multiple solutions. Without loss of generality we also assume that $\mathrm{rank}(A_i)=M$, $\forall i=\overline{1,N}$, i.e., all matrices $A_i$ are full-rank. Table \ref{tab:TableOfNotation} provides quick reference to the variables used in this paper.

\begin{table}
\centering
\vskip 0.1 in
\caption{Notation used in this paper}
\label{tab:TableOfNotation}
\begin{tabular}{ l | l  }
  \hline
  $\tilde X$ & True row sparse matrix of dimension $\mathbb{R}_{+}^{K\times N}$ \\
  $X$ & Unknown row sparse matrix of dimension $\mathbb{R}_{+}^{K\times N}$ \\
  $x_{i}$ & $i$th column of matrix $X$  \\
  $(x^{l})^{\mathrm{T}}$ & $l$th row of matrix X  \\
  $A_{i}$ & $i$th mixing matrix of dimension $\mathbb{R}_{+}^{M\times K}$ \\
  $Y$ & Matrix of measurements / Poisson counts of dimension $\mathbb{N}_{0}^{M\times N}$\\
  \hline
\end{tabular}
\end{table}

We are interested in: 1) finding row sparsest possible matrix $X \in \mathbb{R}_{+}^{K\times N}$ from the observed Poisson measurements $Y$. Ideally, we want to recover initial row sparse matrix $\tilde X$; 2) establishing conditions under which the row sparsity and row sparsity pattern of the recovered matrix $X$ are exactly the same as the row sparsity and row sparsity pattern of the initial row sparse matrix $\tilde X$; and 3) developing an efficient algorithm for recovering the row sparse matrix $X.$

\section{Row Sparsity recovery background}
\label{sec:RowSparsityRecovery}

In this section, we review several approaches for row sparsity recovery. We start by describing the standard least squares and maximum likelihood approaches for recovering the matrix $X$ from the observation matrix $Y$. Then we review methods that introduce sparsity constraints for both LS am ML frameworks. Next we discuss some important issues with those two approaches and motivate our proposed confidence-constrained approach.

\subsection{Unconstrained LS and ML}
\label{ssec:UnconstrainedLSML}

\subsubsection{Unconstrained nonnegative least squares (NNLS)}
\label{sssec:UnconstrainedLS}

The classical least squares method finds the solution that provides the best fit to the observed data. This solution minimizes the sum of squared errors, where an error is the difference between an observed value and the fitted value provided by a model. The nonnegative least squares formulation for our problem is given by:
\begin{equation}
\begin{aligned}
& \underset{X \ge 0}{\text{minimize}}
& & \sum_{i=1}^{N} || A_{i}x_{i}-y_i||_{2}^{2}, \\
\end{aligned}
\label{NNLS}
\end{equation}
where $y_i(j) \sim \mathrm{Poisson}(\lambda_{i}(j)),$ $\lambda_{i}=A_i \tilde x_i,$ $\forall i=\overline{1,N},$ $\forall j=\overline{1,M}.$ Matrix $\tilde X = [\tilde x_1,\tilde x_2,\ldots,\tilde x_N]$ is the true row sparse matrix and matrix $X=[x_1,x_2,\ldots,x_N]$ is the matrix of interest. We require the solution matrix $X$ to be nonnegative due to the Poisson noise assumptions.

\subsubsection{Unconstrained ML}
\label{sssec:UnconstrainedML}
The standard maximum likelihood approach aims to select the values of the model parameters that maximize the probability of the observed data, i.e., the likelihood function. Under the Poisson noise assumptions $y_i(j) \sim \mathrm{Poisson}(\lambda_{i}(j)), \forall i=\overline{1,N}, \forall j=\overline{1,M},$ with $\lambda_{i}=A_i \tilde x_i$, the  probability of observing a particular vector of counts $y_i$ can be written as follows:
\begin{equation}
\label{probpoisson}
P\left(y_i|x_i\right)=\prod_{j=1}^{M}\exp(-\lambda_{i}(j))\frac{\left(\lambda_{i}(j)\right)^{y_{i}(j)}}{y_{i}(j)!},
\end{equation}
where $\lambda_{i} = A_i \tilde x_i,$ and $\lambda_{i} \in \mathbb{R}_{+}^{M\times 1}.$
Using the fact that Poisson distributed observations $y_{i}(j)$ are independent $\forall i=\overline{1,N}, \forall j=\overline{1,M}$, we obtain the negative log-likelihood function $J(X)$:
\begin{equation}
\label{logpoisson}
J(X) = - \log P\left(Y|X\right) = - \log \prod_{i=1}^{N}P\left(y_i|x_i\right) = \sum_{i=1}^{N}\sum_{j=1}^{M}\Big(\lambda_{i}(j)-y_{i}(j)\log\lambda_{i}(j)+\log(y_{i}(j)!)\Big).
\end{equation}
By adding and subtracting the constant term $\sum_{i=1}^{N}\sum_{j=1}^{M}\big(y_{i}(j)\log y_{i}(j)-y_{i}(j)\big)$, we rewrite the negative log-likelihood function $J(X)$ as $J(X) = \sum_{i=1}^{N}I\left(y_{i}||\lambda_{i}\right)+C$, where $C = \sum_{i=1}^{N} \sum_{j=1}^{M} \Big(\log(y_{i}(j)!)-y_{i}(j)\log y_{i}(j)+y_{i}(j)\Big)$ is not a function of $X$, and $I\left(y_{i}||\lambda_{i}\right)$ is the I-divergence, that is a function of $X$ since $\lambda_i = A_i \tilde x_i, \forall i = \overline{1,N}.$

Maximum likelihood approach requires minimization of the negative log-likelihood function $J(X)$ with respect to $X$. Since the function $J(X)$ and the sum of I-divergence terms differ only by a constant term $C$, we can omit the constant term and write the optimization problem in the following form:
\begin{equation}
\begin{aligned}
& \underset{X \ge 0}{\text{minimize}}
& & \sum_{i=1}^{N} I(y_i||A_i x_i).\\
\end{aligned}
\label{UncML}
\end{equation}
We can rewrite (\ref{UncML}) explicitly as follows:
\begin{equation}
\begin{aligned}
& \underset{X \ge 0}{\text{minimize}}
& & \sum_{i=1}^{N} \sum_{j=1}^{M} \Big(y_{i}(j)\log\left(\frac{y_{i}(j)}{e_{j}^{\mathrm{T}}A_{i}x_{i}}\right)+e_{j}^{\mathrm{T}}A_{i}x_{i}-y_{i}(j)\Big).\\
\end{aligned}
\label{UncMLexpl}
\end{equation}
Notice that optimization problem (\ref{UncMLexpl}) is convex.

\subsubsection{Discussion}
\label{sssec:ULSMLdiscussion}

Both the unconstrained NNLS (\ref{NNLS}) and unconstrained ML (\ref{UncML}) problems are well-suited for the case of a set of over-determined equations, i.e., when the number of rows of the $A_i$ matrices is greater than the number of columns. However, the estimation of sparse signals is often examined in the setting of a set of under-determined questions, $M \ll K$. In this case, more than one solution may exist. Additionally, the unconstrained NNLS (\ref{NNLS}) and unconstrained ML (\ref{UncML}) formulations do not incorporate any information on row sparsity of the unknown matrix $X$. Therefore, unconstrained NNLS and unconstrained ML do not force the solution to be row sparse.

\subsection{Regularized LS and ML}
\label{ssec:RLSML}
In this section, we discuss one possible way to enforce row sparsity on the solution matrix in problems (\ref{NNLS}) and (\ref{UncML}). Specifically, we introduce the $R^0$-norm regularization term into both the LS and ML optimization frameworks. We start with the LS framework.

\subsubsection{Regularized NNLS}
\label{sssec:RegularizedLS}
In the regularized NNLS, $R^0$-norm is added to the objective function in (\ref{NNLS}) as a penalty term yielding:
\begin{equation}
\begin{aligned}
& \underset{X \ge 0}{\text{minimize}}
& & \sum_{i=1}^{N} || A_{i}x_{i}-y_i||_{2}^{2}+\nu \|X\|_{R^0},
\end{aligned}
\label{RegNNLS}
\end{equation}
where the unknown parameter $\nu$ defines the importance of the $R^0$-norm, i.e., it controls the trade-off between data fit and row sparsity. Since we can apply the Lagrange multipliers framework to substitute a constraint with a regularization term, we can reformulate (\ref{RegNNLS}) as a constrained LS, as follows:
\begin{equation}
\begin{aligned}
& \underset{X \ge 0}{\text{minimize}}
& & \sum_{i=1}^{N} || A_{i}x_{i}-y_i||_{2}^{2} \\
& \text{subject to}
& & \|X\|_{R^{0}} \le \xi,
\end{aligned}
\label{RegNNLS2}
\end{equation}
where the tuning parameter $\xi \ge 0$ is not known and controls the trade-off between data fit and row sparsity. Clearly, problems (\ref{RegNNLS}) and (\ref{RegNNLS2}) are equivalent in the sense that for every value of parameter $\nu$ in (\ref{RegNNLS}) there is a value of parameter $\xi$ in (\ref{RegNNLS2}) that produces the same solution.

\subsubsection{Regularized ML}
\label{sssec:RegularizedML}
Similar to the regularized NNLS, we add $R^0$-norm to the objective function in (\ref{UncML}) as a penalty term yielding:
\begin{equation}
\begin{aligned}
& \underset{X \ge 0}{\text{minimize}}
& & \sum_{i=1}^{N} I(y_i||A_i x_i)+\nu \|X\|_{R^0},\\
\end{aligned}
\label{RegML}
\end{equation}
where the unknown tuning parameter $\nu$ controls the trade-off between data fit and row sparsity.
Applying similar reasoning, we can reformulate (\ref{RegML}) as a constrained ML problem, as follows:
\begin{equation}
\begin{aligned}
& \underset{X \ge 0}{\text{minimize}}
& & \sum_{i=1}^{N} I(y_i||A_i x_i)\\
& \text{subject to}
& & \|X\|_{R^{0}} \le \xi,
\end{aligned}
\label{RegML2}
\end{equation}
where the tuning parameter $\xi \ge 0$ is not known and controls the trade-off between data fit and row sparsity.

\subsubsection{Discussion}
\label{sssec:RLSMLDiscussion}
The choice of the tuning parameters $\nu$ or $\xi$ is an important challenge associated with regularized NNLS and ML formulations. As we have said earlier, solutions to these problems may exhibit a trade-off between data fit and sparsity. A sparse solution may result in a poor data fit while a solution which provides a good data fit may have many non-zero rows. When there is no information on the noise characteristics there is no criterion for choosing the tuning parameters that guarantees the exact row sparsity and row sparsity pattern recovery of the matrix $X$. The problem is common to many noisy reconstruction algorithms including recovery of sparsity, row-sparsity, and rank. Several approaches for choosing the regularization parameter such as L-Curve and Morozov's discrepancy principle were discussed in \cite{morrey2008multiple, lawson1974solving, wahba1977practical, hansen2006exploiting, rust2008residual, babacan2008parameter, oliveira2009adaptive, gorodnitsky1997energy, hansen1992analysis, rao2003subset, ramirez2012low}.

In the next section, we introduce the confidence-constrained row sparsity minimization for the LS and ML frameworks, and demonstrate how the proposed problem formulations address the issue of selecting tuning parameters for the regularized LS and ML frameworks.

\section{Confidence-constrained row sparsity minimization}
\label{sec:CCRSPMIN}

Recall that our goal is to find row sparsest possible solution matrix $X \in \mathbb{R}_{+}^{K\times N}$ from the observed Poisson measurements $Y \in \mathbb{N}_{0}^{M\times N}$. As we have discussed in Sections \ref{sssec:UnconstrainedLS} and \ref{sssec:UnconstrainedML}, the unconstrained NNLS and ML frameworks aim to find the solution that best fits the observed data, with no enforcement of the row sparsity of the solution. Although, adding $R^0$-norm as constraints in both frameworks in Sections \ref{sssec:RegularizedLS} and \ref{sssec:RegularizedML} allows one to account for row sparsity, we had no criterion for choosing the regularization parameters. Now, we reformulate constrained LS and ML problems (\ref{RegNNLS2}) and (\ref{RegML2}) by switching the roles of the objective function and constraints. For each framework, we propose to use $R^0$-norm of matrix $X$, i.e., $\|X\|_{R^0}$ as the objective function. We propose to use the log-likelihood and squared error functions in inequality type constraints of the forms $\sum_{i=1}^{N} I(y_i||A_i x_i) \le \epsilon_{ML}$ and $\sum_{i=1}^{N} \|A_i x_i-y_i\|_2^2 \le \epsilon_{LS},$ where the parameters $\epsilon_{ML}$ and $\epsilon_{LS}$ control the size of the confidence sets. Confidence set is a high-dimensional generalization of the confidence interval. Using statistical properties of the observed data we derive the model based in-probability bounds on data fit criterion, i.e., $\epsilon_{LS}$ and $\epsilon_{ML}$, which restrict the search space of the problem. One advantage of this approach is the ability to guarantee in probability the perfect reconstruction of row sparsity and row sparsity pattern by searching the solutions inside the confidence set specified by $\epsilon_{LS}$ for the LS framework and $\epsilon_{ML}$ for the ML framework. Now we present the confidence-constrained LS and ML problem formulations:

\begin{tabular}{ p{0.45\textwidth} p{0.45\textwidth} }
LS Confidence-constrained row sparsity minimization (LSCC-RSM)
\begin{equation}
\label{CCLS}
\begin{aligned}
& \underset{X \ge 0}{\text{minimize}}
& & \|X\|_{R^{0}}, \\
& \text{subject to}
& & \sum_{i=1}^{N} \|A_i x_i-y_i\|_2^2 \le \epsilon_{LS};
\end{aligned}
\end{equation}
&
ML Confidence-constrained row sparsity minimization (MLCC-RSM)
\begin{equation}
\label{CCML}
\begin{aligned}
& \underset{X \ge 0}{\text{minimize}}
& & \|X\|_{R^{0}}, \\
& \text{subject to}
& & \sum_{i=1}^{N} I(y_i||A_i x_i) \le \epsilon_{ML}.
\end{aligned}
\end{equation}
\tabularnewline
\end{tabular}

New problem formulations (\ref{CCLS}) and (\ref{CCML}) enforce the row sparsity of the solutions through the objective functions, while fitting to the observation data through the inequality type constraints. 
We are interested in a tuning-free method, i.e., a method which fixes $\epsilon_{LS}$ and $\epsilon_{ML}$ to a specific value that guarantees exact row sparsity and row sparsity pattern recovery. Similar to our approach in \cite{chunikhina2012tuning}, using statistical characteristics of the observed data we select the parameters $\epsilon_{LS}$ and $\epsilon_{ML}$ to control the radiuses of the corresponding confidence sets and guarantee exact row sparsity and row sparsity pattern recovery.

\section{Exact row sparsity and row sparsity pattern recovery: Theoretical guarantees}
\label{sec:TheorGuarantees}

In this section, we provide theoretical guarantees for exact row sparsity and row sparsity pattern recovery. We start by presenting two propositions that provide lower and upper bounds on row sparsity of the solution matrix; then we continue with theorems for exact recovery in each framework. Complete proofs for all propositions and theorems are provided in the Appendix section of this paper.

\begin{Def}
\label{confSets}
For a given row sparse matrix $\tilde X\in\mathbb{R}^{K\times N}_{+}$, mixing matrices $A_i \in \mathbb{R}^{M\times K}_{+}$, $\forall i = \overline{1,N}$, and measurement matrix $Y\in \mathbb{N}^{M\times N}_{0}$, such that $y_i(j) \sim \mathrm{Poisson}(\lambda_i(j))$, $\lambda_i = A_{i}\tilde x_{i}$, $\forall i=\overline{1,N}$, define

\begin{itemize}
  \item $S_{\epsilon_{LS}}$ to be the set of all possible matrices $X\in\mathbb{R}^{K\times N}_{+}$ that satisfy LSCC-RSM constraint inequality:
      \begin{equation}
      \label{LSconfset}
      S_{\epsilon_{LS}} = \left \{ X : \sum_{i=1}^{N} \|A_{i}x_{i}-y_{i}\|_{2}^{2} \le \epsilon_{LS}\right \}.
      \end{equation}
      We call set $S_{\epsilon_{LS}}$ - the least squares confidence set.
  \item $S_{\epsilon_{ML}}$ to be the set of all possible matrices $X\in\mathbb{R}^{K\times N}_{+}$ that satisfy MLCC-RSM constraint inequality:
      \begin{equation}
      \label{MLconfset}
      S_{\epsilon_{ML}} = \left \{ X : \sum_{i=1}^{N} I(y_{i}||\lambda_{i}) \le \epsilon_{ML}\right \},
      \end{equation}
      We call set $S_{\epsilon_{ML}}$ - the maximum likelihood confidence set.
\end{itemize}
\end{Def}

\begin{prop}(Upper bound on row sparsity)
\label{propUpperBound}

Let $\tilde X\in\mathbb{R}^{K\times N}_{+}$ be the true row sparse matrix. Let $S$ be a confidence set that is either the least squares or the maximum likelihood confidence set.
If the true row sparse matrix $\tilde X$ is in $S$, i.e., $\tilde X \in S$ then any $X^* \in \arg \min_{X\in S}\|X\|_{R^0}$ satisfies $\|X^*\|_{R^0} \le \|\tilde X\|_{R^0}.$
\end{prop}

This proposition suggests that if a confidence set $S$ is sufficiently large to contain the true matrix $\tilde X$ then the row sparsity of the solution $X^*$ to the optimization problem $\min_{X\in S}\|X\|_{R^0}$ is less than or equal to the row sparsity of the true row sparse matrix $\tilde X$. Next we present two propositions, where for each framework LS and ML we obtain $\epsilon_{LS}$ and $\epsilon_{ML}$, i.e., radiuses of the confidence sets $S_{\epsilon_{LS}}$ and $S_{\epsilon_{ML}}$ such that with high probability we guarantee $\tilde X \in S_{\epsilon_{LS}}$ and $\tilde X \in S_{\epsilon_{ML}}$ respectively.

\begin{prop} (Parameter-free radius of confidence set for least squares framework)
\label{propRadiusLS}

Let $\tilde X\in\mathbb{R}^{K\times N}_{+}$ be the true row sparse matrix. Let the mixing matrices $A_i \in \mathbb{R}^{M\times K}_{+}$, $\forall i = \overline{1,N}$ be given. Let measurement matrix $Y\in \mathbb{N}^{M\times N}_{0}$ be obtained as following: $y_i(j)\sim \mathrm{Poisson}(\lambda_{i}(j))$, where $\lambda_{i}=A_{i}\tilde x_{i} $, $\forall i=\overline{1,N},j=\overline{1,M}$. Let the least squares confidence set be $S_{\epsilon_{LS}}$.
Let $p \in (0,1)$  be given. Let $\psi = \sum_{i=1}^{N}\sum_{j=1}^{M} y_{i}(j)$ and $k = \sqrt{\frac{2}{p}-1}$.
Set $\epsilon_{LS}$ according to
\begin{equation}
\label{formula_epsilon_LS}
\epsilon_{LS}(p) = \psi+\frac{k^{2}}{2}+\sqrt{\psi k^2+\frac{k^{4}}{4}}+k \sqrt{2\psi^{2}+\psi(4k^{2}+1)+k^{4}+\frac{k^{2}}{2}+
\left(4\psi + 2k^2+1\right)\sqrt{\psi k^2+\frac{k^{4}}{4}}}.
\end{equation}
Then with probability at least $1-p$, $ \tilde X \in S_{\epsilon_{LS}}. $
\end{prop}

\begin{prop}(Parameter-free radius of confidence set for maximum likelihood framework)
\label{propRadiusML}

Let $\tilde X\in\mathbb{R}^{K\times N}_{+}$ be the true row sparse matrix. Let the mixing matrices $A_i \in \mathbb{R}^{M\times K}_{+}$, $\forall i = \overline{1,N}$ be given. Let  measurement matrix $Y\in \mathbb{N}^{M\times N}_{0}$ be obtained as following: $y_i(j)\sim \mathrm{Poisson}(\lambda_{i}(j))$, where $\lambda_{i}=A_{i}\tilde x_{i} $, $\forall i=\overline{1,N},j=\overline{1,M}$. Let the the maximum likelihood confidence set be $S_{\epsilon_{ML}}$.
Let $p \in (0,1)$  be given. Set $C_{\mu} \approx 0.5801$ and $C_{\sigma^2} \approx 0.5178$. Set $\epsilon_{ML}$ according to
\begin{equation}
\label{formula_epsilon_ML}
\epsilon_{ML}(p) = C_{\mu} MN +\sqrt{\frac{1}{p}-1}\sqrt{C_{\sigma^2} MN}.
\end{equation}
Then with probability at least $1-p$, $ \tilde X \in S_{\epsilon_{ML}}. $
\end{prop}

\begin{cor}
\label{cor}

Let $\tilde X\in\mathbb{R}^{K\times N}_{+}$ be the true row sparse matrix. Let mixing matrices $A_i \in \mathbb{R}^{M\times K}_{+}$, $\forall i = \overline{1,N}$ be given. Let measurement matrix $Y\in \mathbb{N}^{M\times N}_{0}$ be obtained as following: $y_i(j)\sim \mathrm{Poisson}(\lambda_{i}(j))$, where $\lambda_{i}=A_{i}\tilde x_{i} $, $\forall i=\overline{1,N},j=\overline{1,M}$. Let $p \in (0,1)$  be given.
\begin{itemize}
  \item If $X^{LS}$ is a solution matrix to LSCC-RSM optimization problem (\ref{CCLS}) where $\epsilon_{LS}$ is given by (\ref{formula_epsilon_LS}) then with probability at least $1-p$, $X^{LS}$ satisfies $\|X^{LS}\|_{R^0} \le \|\tilde X\|_{R^0}.$
  \item If $X^{ML}$ is a solution matrix to MLCC-RSM optimization problem (\ref{CCML}) where $\epsilon_{ML}$ is given by (\ref{formula_epsilon_ML}) then with probability at least $1-p$, $X^{ML}$ satisfies $\|X^{ML}\|_{R^0} \le \|\tilde X\|_{R^0}.$
\end{itemize}
\end{cor}

This corollary suggests that the LSCC-RSM and MLCC-RSM optimization problems (\ref{CCLS}) and (\ref{CCML}) produce solutions with row sparsity which is less than or equal to the row sparsity of the true row sparse matrix $\tilde X$ if the corresponding confidence sets $S_{\epsilon_{LS}}$ and $S_{\epsilon_{ML}}$ are sufficiently large to contain the original matrix $\tilde X$. Proof of Corollary \ref{cor} follows directly from the proofs of Propositions \ref{propUpperBound}, \ref{propRadiusLS}, and \ref{propRadiusML}.

\begin{prop}(Lower bound on row sparsity)
\label{propLowerBound}

Let $\tilde X\in\mathbb{R}^{K\times N}_{+}$ be the true row sparse matrix. Define $ \gamma(X) = \min_{l=\overline{1,K}} \{ \|e_l^{\mathrm{T}} X\|_2 :  \|e_l^{\mathrm{T}} X\|_2\neq 0 \}. $ Then for any matrix $X$ satisfying
$\|\tilde{X}-X\|_F < \gamma(\tilde X)$, we have $\|X\|_{R^0} \ge \|\tilde X\|_{R^0}$.
\end{prop}

This proposition suggests that all matrices that are inside the $\gamma(\tilde X)$-Frobenius ball neighborhood of $\tilde X$, i.e., $X \in \mathrm{B}_{\gamma(\tilde X)}^{F}(\tilde X)$ have row sparsity greater or equal to the row sparsity of the $\tilde X.$ Our intuition is that since matrix $\tilde X$ has sufficiently large row norms then small changes to matrix $\tilde X$ cannot set its rows to zero and therefore cannot lower the row sparsity of $\tilde X$.

Next, we want to combine the results of Propositions \ref{propUpperBound} and \ref{propLowerBound} for both frameworks. For example, consider the LSCC-RSM optimization problem (similar reasoning applies to the MLCC-RSM optimization problem). If the true row sparse matrix $\tilde X$ is inside the confidence set $S_{\epsilon_{LS}}$ and the confidence set $S_{\epsilon_{LS}}$ is a subset of the $\gamma_{LS}(\tilde X)$-Frobenius ball neighborhood of $\tilde X$, i.e., $\tilde X \in S_{\epsilon_{LS}} \subseteq \mathrm{B}_{\gamma_{LS}(\tilde X)}^{F}(\tilde X)$ then the solution to the LSCC-RSM optimization problem $X^{LS}$ must satisfy inequality $\|X^{LS}\|_{R^0} \le \|\tilde X\|_{R^0}$ by Proposition \ref{propUpperBound} and the inequality $\|X^{LS}\|_{R^0} \ge \|\tilde X\|_{R^0}$ by Proposition \ref{propLowerBound}, therefore guaranteeing $\|X^{LS}\|_{R^0} = \|\tilde X\|_{R^0}$.

We present two theorems which set the conditions for $S_{\epsilon_{LS}} \subseteq \mathrm{B}_{\gamma_{LS}(\tilde X)}^{F}(\tilde X)$ for LS framework and conditions for $S_{\epsilon_{ML}} \subseteq \mathrm{B}_{\gamma_{ML}(\tilde X)}^{F}(\tilde X)$ for ML framework assuring that not only row sparsity of the corresponding solution matrix is the same as the row sparsity of the true matrix $\tilde X$, but also that the row sparsity pattern of the solution matrix is the same as the row sparsity pattern of the true matrix $\tilde X$.

\begin{thm}(Exact recovery for LSCC-RSM)
\label{thmLS}

Let $\tilde X\in\mathbb{R}^{K\times N}_{+}$ be the true $s$-row sparse matrix. Let the mixing matrices $A_i \in \mathbb{R}^{M\times K}_{+}$, $\forall i = \overline{1,N}$ be given such that $\forall i = \overline{1,N}$ matrices $A_i$ satisfy $2s$-restricted isometry property with $\delta_{2s} <1 $. Let measurement matrix $Y\in \mathbb{N}^{M\times N}_{0}$ be obtained as follows: $y_i(j)\sim \mathrm{Poisson}(\lambda_{i}(j))$, where $\lambda_{i}=A_{i}\tilde x_{i} $, $\forall i=\overline{1,N}$. Let $S_{\epsilon_{LS}}$ be the least squares confidence set given by (\ref{LSconfset}), and $\epsilon_{LS}$ chosen according to Proposition \ref{propRadiusLS}, so that $\tilde X \in S_{\epsilon_{LS}}$ with probability at least $1-p$, $p \in(0,1)$.
If $\gamma_{LS}(\tilde X) \ge \frac{2\sqrt{{\epsilon_{LS}}}}{1-\delta_{2s}},$
then $S_{\epsilon_{LS}} \subseteq \mathrm{B}_{\gamma_{LS}(\tilde X)}^{F}(\tilde X)$ and the solution $X^{LS}$ to the LSCC-RSM problem (\ref{CCLS}) satisfies both $\|X^{LS}\|_{R^0}= \|\tilde X\|_{R^0}$ and $\mathrm{RSupp}(X^{LS}) = \mathrm{RSupp}(\tilde X)$ with probability at least $1-p$.
\end{thm}

\begin{thm}(Exact recovery for MLCC-RSM)
\label{thmML}

Let $\tilde X\in\mathbb{R}^{K\times N}_{+}$ be the true $s$-row sparse matrix. Let the mixing matrices $A_i \in \mathbb{R}^{M\times K}_{+}$, $\forall i = \overline{1,N}$ be given. Assume matrices $A_i$ satisfy $2s$-restricted isometry property with $\delta_{2s} <1 $. Let measurement matrix $Y\in \mathbb{N}^{M\times N}_{0}$ be obtained as follows: $y_i(j)\sim \mathrm{Poisson}(\lambda_{i}(j))$, where $\lambda_{i}=A_{i}\tilde x_{i} $, $\forall i=\overline{1,N}$. Let $S_{\epsilon_{ML}}$ be the maximum likelihood confidence set given by (\ref{MLconfset}), and $\epsilon_{ML}$ chosen according to proposition (\ref{propRadiusML}), so that $\tilde X \in S_{\epsilon_{ML}}$ with probability at least $1-p$, $p \in(0,1)$. Let $G(z) = \frac{(1+\sqrt{2z})(z+\log(1+\sqrt{2z})-\sqrt{2z})}{\sqrt{2z}}.$
If $$\gamma_{ML}(\tilde X) \ge  \frac{2\sqrt{ \epsilon_{ML}\sum_{i=1}^{N}\left(2\sqrt{2 \|y_i\|_1}+\frac{\|y_i\|_1}{\sqrt{\epsilon_{ML}}}G\left(\frac{2\epsilon_{ML}}{\|y_i\|_1}\right)\right)^2}}{1-\delta_{2s}}$$
then $S_{\epsilon_{ML}} \subseteq \mathrm{B}_{\gamma_{ML}(\tilde X)}^{F}(\tilde X)$ and the solution $X^{ML}$ to the MLCC-RSM optimization problem (\ref{CCML}) satisfies both $\|X^{ML}\|_{R^0} = \|\tilde X\|_{R^0}$ and $\mathrm{RSupp}(X^{ML}) = \mathrm{RSupp}(\tilde X)$ with probability at least $1-p$.
\end{thm}

Theorems \ref{thmLS} and \ref{thmML} suggest for each framework LS and ML that if the true row sparse matrix $\tilde X$ is in the corresponding confidence set $S_{\epsilon_{LS}}$ or $S_{\epsilon_{ML}}$ and each one of $\tilde X$ $s$ non-zero rows has sufficient large $l_2$-norm, then the corresponding solutions of LSCC-RSM and MLCC-RSM optimization problems will have the same row sparsity and row sparsity pattern as that of $\tilde X$.

\section{Convex relaxation}
\label{sec:ConvexRelaxation}

In practice, solutions to the problems (\ref{CCLS}) and (\ref{CCML}) are computationally infeasible due to the combinatorial nature of the objective function $\|X\|_{R^0}$ \cite{natarajan1995sparse}. One of the effective approaches proposed to overcome this issue is to exchange the original objective function $\|X\|_{R^0}$ with the mixed $\ell_{p,q}$ norm of $X$ defined as $\| X \|_{p,q}=\left(\sum_{l=1}^{K}\|(x^l)^{\mathrm{T}}\|_q^p\right)^{1/p},$ where the parameters $p$ and $q$ are such that $0<p<\infty$ and $0<q\le \infty$.

A class of optimization problems with $\| X \|_{p,q}$ objective function, i.e., problems of the form:
\begin{equation*}\label{cvx}
\begin{aligned}
& \underset{X}{\text{minimize}}
& & \|X\|_{p,q} \\
& \text{subject to}
& & Y=AX,
\end{aligned}
\end{equation*}
was studied in \cite{cotter2005sparse, tropp2006algorithms1, tropp2006algorithms2, chen2006theoretical, rao1998basis, malioutov2005sparse, eldar2008robust}. In particular, the choice of parameters $p = 1$, $q = 2$ was considered in \cite{malioutov2005sparse}, \cite{eldar2008robust}; $p = 1$, $q = \infty$ in \cite{tropp2006algorithms1}, \cite{tropp2006algorithms2}; $p = 1$, $q \ge 1$ in \cite{chen2006theoretical}; $p = 2$, $q \le 1$ in \cite{cotter2005sparse}; $p \le 1$, $q = 2$ in \cite{rao1998basis, cotter2005sparse, malioutov2005sparse}; $p = 2$, $q = 0$ in \cite{hyder2009robust}. Some discussion and comparison of different methods was given in \cite{berg2009joint, tropp2006algorithms2, hyder2009robust}.

We propose to transform our original optimization problems (\ref{CCLS}) and (\ref{CCML}) by exchanging the objective function $\|X\|_{R^0}$ with $\|X\|_{1,2} = \sum_{l=1}^{K} \|(x^l)^{\mathrm{T}}\|_2$, i.e., assigning $p = 1$, $q = 2$. The assignment $p = 1$, $q = 2$ is intuitive. To enforce row sparsity of matrix $X$, we first obtain a vector by computing $\ell_2$-norm of all rows of $X$. Then we enforce sparsity of this new vector by computing its $\ell_1$-norm. Enforcing $\ell_1$-norm of vector of the row amplitudes to be sparse is equivalent to enforcing the entire rows of $X$ to be zero.

Consequently, we propose a following convex relaxation to the optimization problems (\ref{CCLS}) and (\ref{CCML}):

\begin{tabular}{ p{0.45\textwidth} p{0.45\textwidth} }
LS confidence-constrained $\ell_{1,2}$ minimization (LSCC-L12M)
\begin{equation}
\label{CCLScvx}
\begin{aligned}
& \underset{X \ge 0}{\text{minimize}}
& & \|X\|_{1,2} \\
& \text{subject to}
& & \sum_{i=1}^{N} \|A_i x_i-y_i\|_2^2 \le \epsilon_{LS},
\end{aligned}
\end{equation}
&
ML confidence-constrained $\ell_{1,2}$ minimization (MLCC-L12M)
\begin{equation}
\label{CCMLcvx}
\begin{aligned}
& \underset{X \ge 0}{\text{minimize}}
& & \|X\|_{1,2} \\
& \text{subject to}
& & \sum_{i=1}^{N} I(y_i||A_i x_i) \le \epsilon_{ML}.
\end{aligned}
\end{equation}
\tabularnewline
\end{tabular}

\section{Solution: Gradient approach}
\label{sec:gradappr}

In this section, we propose to solve convex optimization problems (\ref{CCLScvx}) and (\ref{CCMLcvx}) by minimizing corresponding Lagrangian function and using binary search to find the Lagrange multiplier. To describe the intuition for our approach, consider the following optimization problem:
\begin{equation}
\label{generalProblem}
\begin{aligned}
& \underset{X \in \mathcal{X}}{\text{minimize}}
& & f(X) \\
& \text{subject to}
& & g(X) \le 0,
\end{aligned}
\end{equation}
where function $f(X)$ is strictly convex and function $g(X)$ is convex.

The Lagrangian for problem (\ref{generalProblem}) is defined as $L(X;\lambda) = f(X)+\lambda g(X)$. The solution is given by $(X^*;\lambda^*) = \arg \max_{\lambda}\min_{X \in \mathcal{X}}L(X;\lambda)$. 
To determine the proper value of the regularization parameter $\lambda$ recall the Karush-Kuhn-Tucker (KKT) optimality conditions. The complementary slackness condition, $\lambda g(X) = 0$ gives us two possible scenarios. First, $g(X) <0$ is possible only if $\lambda = 0$. However, in this case optimal solution will be $X$ that minimizes $f(X)$ without any attention to constraints, i.e., this case leads to a trivial solution (for both LSCC-L12M and MLCC-L12M problems trivial solution is $X^* = 0$). We focus on the second case where $\lambda \ne 0$, enforcing $g(X^*)=0$. Therefore, we are looking for an optimal parameter $\lambda^*$ with an optimal $X^* = X^*(\lambda)$, where $g(X^*(\lambda))=0$ at $\lambda = \lambda^*$. In order to find such optimal $\lambda^*$ we propose a simple binary search. In Algorithm \ref{algOuterIter} we describe the outer search iterations for finding optimal regularization parameter $\lambda^*$.

\begin{algorithm}
\caption{Binary search for regularization parameter $\lambda$}\label{algorithm:regularization}
\label{algOuterIter}
{{\bf Input}: Lagrangian range: $ (\lambda_{min}, \lambda_{max}); \lambda_{min}, \lambda_{max} \in \mathbb{R}$ \\
precision: $\eta_1>0$ \\
{\bf Output}: $\lambda^*$, $X^*$}
\begin{algorithmic}[1]
\STATE {$MaxIteration \gets \lceil \log_2 \left(\frac{\lambda_{max}-\lambda_{min}}{2\eta_1}\right)\rceil$}
\FOR{$r = 1$ \TO $MaxIteration$}
\STATE {$\lambda^r \gets (\lambda_{min} + \lambda_{max})/2$}
\STATE {Find $X(\lambda^r) \gets \arg\min_{X \in \mathcal{X}}L(X;\lambda^r)$ using Algorithm \ref{backtracking}}
\IF {$g(X(\lambda^r))>0$}
\STATE {$\lambda_{min} \gets \lambda^r$}
\ELSE {
\STATE {$\lambda_{max} \gets \lambda^r$}
}
\ENDIF
\ENDFOR
\STATE {$\lambda^* \gets \lambda^r$}
\STATE {$X^* \gets X(\lambda^*)$}
\end{algorithmic}
\end{algorithm}

Algorithm \ref{algOuterIter} suggests that we know the interval for binary search, i.e., $[\lambda_{min},\lambda_{max}]$. To find this interval we simply find two values of $\lambda$, say $\lambda_1$ and $\lambda_2$ for which constraints function $g(X^*(\lambda))$ satisfies the following inequality: $g(X^*(\lambda_1))>0>g(X^*(\lambda_2))$. Since the function $f(X)$ is strictly convex, the function $L(X;\lambda)$ is strictly convex and hence has a unique minimum and unique minimizer; this makes function $g(X^*(\lambda))$ to be a continuous function of $\lambda$.
Hence by Intermediate Value Theorem \cite[p.~157]{johnsonbaugh2012foundations} there exist such $\lambda^* \in (\lambda_1,\lambda_2)$ for which $g(X^*(\lambda^*))=0$. Moreover, using a binary search method we can find the optimal $\lambda^*$ within $\BigO{\log \left(\frac{1}{\eta_1}\right)}$ steps.

Now, given a particular $\lambda $ we can find solution $X^*(\lambda) = \arg\min_{X}L(X;\lambda)$ by minimizing Lagrangian using projected subgradient approach, i.e., we first take steps proportional to the negative of the subgradient of the function at the current point, then we project the solution to the nonnegative orthant because of nonnegativity constraints on the solution matrix $X$. To find an efficient gradient step size we use well-known backtracking linesearch strategy \cite[p.~464]{boyd2004convex}. Algorithm \ref{backtracking} provides overview of this method.

\begin{algorithm}
\caption{Gradient descent with backtracking linesearch strategy}
\label{backtracking}
{{\bf Input}: $ \lambda^*, c_1, c_2$ \\
precision: $\eta_2>0$ \\
{\bf Output}: $X^*(\lambda^*)$}
\begin{algorithmic}[1]
\STATE $\alpha^0 \gets 1$
\STATE $r \gets 1$
\STATE $X^r \gets Random(K,N)$
\WHILE { $|g(X^r(\lambda^*))| >\eta_2 $ }
\STATE $\alpha^{r} \gets c_1\alpha^{r-1}$
\WHILE {$L\left(X^{r}-\alpha^{r} \nabla L(X^{r});\lambda^*\right)>L\left(X^{r};\lambda^*\right)-\alpha^{r} \|\nabla L\left(X^{r};\lambda^*\right)\|_2^2 $}
\STATE {$\alpha^{r} \gets c_2\alpha^{r}$}
\ENDWHILE
\STATE {$X^{r+1} \gets (X^{r} - \alpha^{r} \nabla L\left(X^{r};\lambda^*\right))_{+}$ }
\STATE {$r \gets r+1$}
\ENDWHILE
\STATE {$X^*(\lambda^*) \gets X^{r} $}
\end{algorithmic}
\end{algorithm}

The values of constants $c_1$ and $c_2$ in Algorithm \ref{backtracking} should be chosen such that $c_1>1$ and $c_2<1$ (e.g. $c_1 = 10$ and $c_2 = 0.5$).

We proceed with the high level derivations for convex LSCC-L12M (\ref{CCLScvx}) and MLCC-L12M (\ref{CCMLcvx}) problems.

\subsection{Gradient approach to LSCC-L12M}
\label{ssec:GradientCCLS}

The Lagrangian for convex LSCC-L12M problem (\ref{CCLScvx}) is given as:
\begin{eqnarray}
\label{CCLS convex Lagrangian}
L_{LS}(X;\lambda_{LS}) & = & \|X\|_{1,2}+\lambda_{LS} \left(\sum_{i=1}^{N} \|A_i x_i-y_i\|_2^2 - \epsilon_{LS}\right) \\ \nonumber
 & = & \sum_{l=1}^{K} \|\left(x^{l}\right)^{\mathrm{T}}\|_2+\lambda_{LS} \left(\sum_{i=1}^{N} \sum_{j=1}^{M}\left[\sum_{l=1}^{K} A_i(j,l)X(l,i)-y_i(j)\right]^2 - \epsilon_{LS}\right) \\ \nonumber
\end{eqnarray}
Differentiating the Lagrangian in (\ref{CCLS convex Lagrangian}) with respect to matrix entry $X(t,s)>0$, $\forall t = \overline{1,K}, s = \overline{1,N}$, we obtain the gradient matrix $\nabla L_{LS}(X;\lambda_{LS}) = \frac{\partial L_{LS}(X;\lambda_{LS})}{\partial X}$ :

\begin{equation}
\label{CCLSgradient}
\nabla L_{LS}(X;\lambda_{LS})\Big|_{(t,s)}  = \frac{\partial L_{LS}(X;\lambda_{LS})}{\partial X(t,s)} = \frac{X(t,s)}{\|\left(x^{t}\right)^{\mathrm{T}}\|_2} + 2\lambda_{LS} \left( A_s x_s-y_s\right)^{\mathrm{T}}A_s e_t.
\end{equation}

Notice, that if at least one row of $X$ is zero then term $\frac{X(t,s)}{\|\left(x^{t}\right)^{\mathrm{T}}\|_2}$ does not exist. To overcome this issue, we use gradient matrix $\nabla L_{LS}$ to define subgradient matrix $\nabla \tilde L_{LS}$ such that term $\frac{X(t,s)}{\|\left(x^{t}\right)^{\mathrm{T}}\|_2}$ is equal to zero at the places that correspond to the zero rows of $X$, i.e.,
\begin{equation}
\label{CCLSsubgradient}
\nabla \tilde L_{LS}(X;\lambda_{LS})\Big|_{(t,s)} =
\begin{cases}
  \frac{X(t,s)}{\|\left(x^{t}\right)^{\mathrm{T}}\|_2} + 2\lambda_{LS} \Big( A_s x_s-y_s\Big)^{\mathrm{T}}A_s e_t,  & \|\left(x^{t}\right)^{\mathrm{T}}\|_2 \ne 0 \\
  0+ 2\lambda_{LS} \Big( A_s x_s-y_s\Big)^{\mathrm{T}}A_s e_t, & \|\left(x^{t}\right)^{\mathrm{T}}\|_2=0.
\end{cases}
\end{equation}

Now, the gradient descent approach implies the following update of the solution matrix $X_{LS}$ on iteration $r+1$:
\begin{equation}
\label{CCLScvxiter}
X^{r+1}_{LS} = \left(X^{r}_{LS} - \alpha_{LS}^r \nabla \tilde L_{LS}\right)_{+},
\end{equation}
where $\alpha_{LS}^r$ is the gradient step size, chosen via line search method described in Algorithm \ref{backtracking}. Notice that we project matrix $X$ on positive orthant because of nonnegativity conditions on matrix $X$.

\subsection{Gradient approach to MLCC-L12M}
\label{ssec:GradientCCML}

The Lagrangian for convex MLCC-L12M problem (\ref{CCMLcvx}) is given as:
\begin{eqnarray}
\label{CCML convex Lagrangian}
\nonumber
L_{ML}(X;\lambda_{ML}) & = & \|X\|_{1,2}+\lambda_{ML} \left(\sum_{i=1}^{N} I(y_i||A_i x_i) - \epsilon_{ML}\right) \\
 & = & \!\!\sum_{l=1}^{K} \|(x^{l})^{\mathrm{T}}\|_2 + \!\!\lambda_{ML} \Big(\sum_{i=1}^{N} \Big[1^{\mathrm{T}} A_i x_i -\!\!\!\sum_{j=1}^{M}y_i(j)\log(e_j^{\mathrm{T}} A_i x_i)+C_i\Big]-\!\epsilon_{ML}\Big),
\end{eqnarray}
where $C_i = \sum_{j=1}^{M}\left(y_i(j)\log(y_i(j))-y_i(j)\right)$ is a constant in $X$ term $\forall i = \overline{1,N}$.

Now, we differentiate Lagrangian (\ref{CCML convex Lagrangian}) with respect to matrix entry $X(t,s)>0$, $\forall t = \overline{1,K}, s = \overline{1,N}$ to obtain the gradient matrix $\nabla L_{ML}(X;\lambda_{ML}) = \frac{\partial L_{ML}(X;\lambda_{ML})}{\partial X}$ :
\begin{eqnarray}
\label{CCMLgradient}
\nonumber
\nabla L_{ML}(X;\lambda_{ML})\Big|_{(t,s)} & = & \frac{\partial L_{ML}(X;\lambda_{ML})}{\partial X(t,s)} \\
& = & \frac{X(t,s)}{\|\left(x^{t}\right)^{\mathrm{T}}\|_2} + \lambda_{ML} \left(\sum_{j=1}^{M} A_s(j,t)-\sum_{j=1}^{M}\left[\frac{y_s(j)A_s(j,t)}{\sum_{l=1}^{K}A_s(j,l)X(l,s)}\right] \right).
\end{eqnarray}

Notice, that if at least one row of $X$ is zero then term $\frac{X(t,s)}{\|\left(x^{t}\right)^{\mathrm{T}}\|_2}$ does not exist. Similar to LSCC-L12M case, we use gradient matrix $\nabla L_{ML}$ to define subgradient matrix $\nabla \tilde L_{ML}$ such that term $\frac{X(t,s)}{\|\left(x^{t}\right)^{\mathrm{T}}\|_2}$ is equal to zero at the places that correspond to the zero rows of $X$, i.e.,
\begin{equation}
\label{CCMLsubgradient}
\nabla \tilde L_{ML}(X;\lambda_{ML})\Big|_{(t,s)} \!\!\!\!=\!
\begin{cases}
  \frac{X(t,s)}{\|(x^{t})^{\mathrm{T}}\|_2} + \lambda_{ML} \Big(\sum_{j=1}^{M} A_s(j,t)-\sum_{j=1}^{M}\Big[\frac{y_s(j)A_s(j,t)}{\sum_{l=1}^{K}A_s(j,l)X(l,s)}\Big] \Big),  & \!\!\! \|(x^{t})^{\mathrm{T}}\|_2 \ne 0 \\
  0 + \lambda_{ML} \Big(\sum_{j=1}^{M} A_s(j,t)-\sum_{j=1}^{M}\Big[\frac{y_s(j)A_s(j,t)}{\sum_{l=1}^{K}A_s(j,l)X(l,s)}\Big] \Big), & \!\!\! \|(x^{t})^{\mathrm{T}}\|_2=0.
\end{cases}
\end{equation}

Now, taking into account nonnegativity constraints on solution matrix $X$, the gradient descent approach implies the following update of the solution matrix $X_{ML}$ on iteration $r+1$:
\begin{equation}
\label{CCMLcvxiter}
X^{r+1}_{ML} = \left(X^{r}_{ML} - \alpha_{ML}^r \nabla \tilde L_{ML}\right)_{+},
\end{equation}
where $\alpha_{ML}^r$ is the gradient step size, chosen via line search method described in Algorithm \ref{backtracking}.

In the next section, we present numerical results on the row sparsity and row sparsity pattern recovery performance of our proposed algorithms.

\section{Simulations}
\label{sec:Simulations}

Recall that in Section \ref{sec:TheorGuarantees} we presented theoretical guarantees on row sparsity and row sparsity pattern recovery for non-convex problems LSCC-RSM (\ref{CCLS}) and MLCC-RSM (\ref{CCML}). In this section, we propose two algorithms for convex problems LSCC-L12M (\ref{CCLScvx}) and MLCC-L12M (\ref{CCMLcvx}), and evaluate their row sparsity and row sparsity pattern recovery performance. We show that the solution to the relaxed problems LSCC-L12M (\ref{CCLScvx}) and MLCC-L12M (\ref{CCMLcvx}) exhibits similar behavior to the solutions to the non-convex problems LSCC-RSM (\ref{CCLS}) and MLCC-RSM (\ref{CCML}), i.e., empirical results of minimizing $\ell_{1,2}$ norm are consistent with theoretical results obtained for row sparsity minimization. 
We provide the following: 1) sensitivity analysis of row sparsity recovery accuracy as a function of $\epsilon$, and 2) probability of correct row sparsity and row sparsity pattern recovery analysis applied to a synthetic data. The numerical experiments were run in MATLAB 7.13.0.564 on a HP desktop with an Intel Core 2 Quad 2.66GHz CPU and 3.49GB of memory.

\subsection{Synthetic data generation}
\label{ssec:datageneration}

We use synthetic data that follows Poisson noise model (\ref{Poisson}). The data is generated as follows. We set $N = 10$ , $M = 30$, $K = 50$. To produce mixing matrices $A_i \in \mathbb{R}^{M\times K}_{+}$, $i=\overline{1,N}$, we generate each entry as an independent Bernoulli trial with probability of success equal to $0.7$, and normalize each column so that it has unit Euclidian norm. To generate the true row sparse matrix $\tilde X \in \mathbb{R}^{K\times N}$ we first set the desired row sparsity of $\tilde X$ to a fixed number. Here we set it to three, i.e., $\|\tilde X\|_{R^0} = 3$. Then we randomly select $3$ rows of $\tilde X$, and fill them with absolute values of i.i.d. $\mathcal{N}(0,1)$ random variables. The remaining $K-3$ rows of are $\tilde X$ filled with zeros. Then we simply check that every matrix $A_i$ satisfies RIP and set RIP constant $\delta_{2s} = \delta_6 <1$ to be the maximum over all RIP constants of matrices $A_i$, $i=\overline{1,N}$. To control the amount of noise we introduce the intensity parameter $\theta$. 
To vary the amount of noise we simply set $\tilde X \gets \theta \tilde X$. Finally, the observation matrix $Y \in \mathbb{N}^{M\times N}_{0}$ is defined columnwise by setting $y_i(j)\sim \mathrm{Poisson}(\lambda_i(j))$, $\lambda_i = A_i \tilde x_i$, $\forall i = \overline{1,N}$.

\subsection{Row sparsity versus different values of $\epsilon$ }
\label{ssec:rspVSeps}
In this section we illustrate the effect of the choice of $\epsilon$ on row sparsity and row sparsity pattern recovery. First, we describe the experiment and its results for the LSCC-L12M problem.

\begin{figure}
  \begin{center}
  \subfigure[]{
  \label{fig6a}
  \includegraphics[scale=0.44]{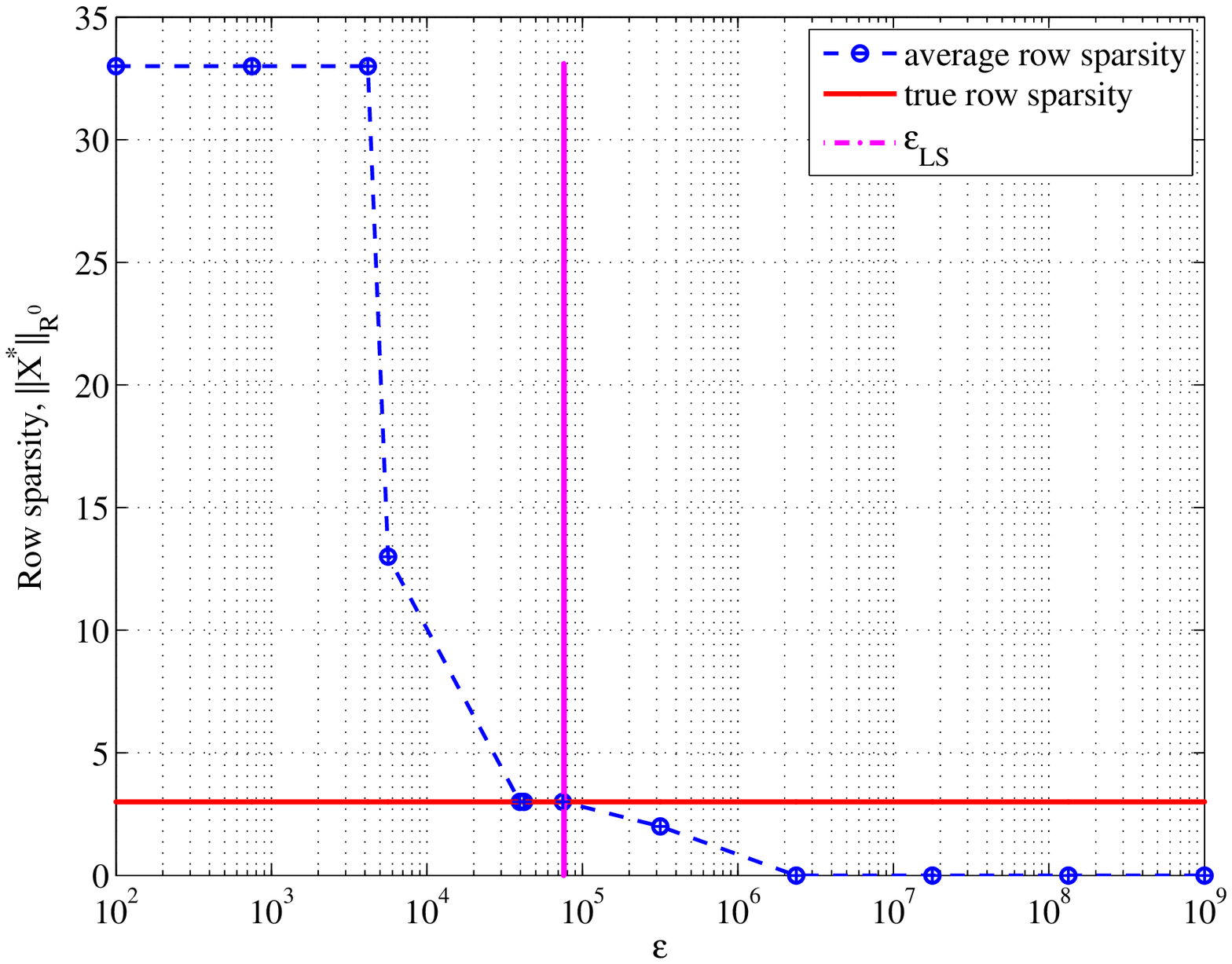}
  }
  \subfigure[]{
  \label{fig6b}
  \includegraphics[scale=0.44]{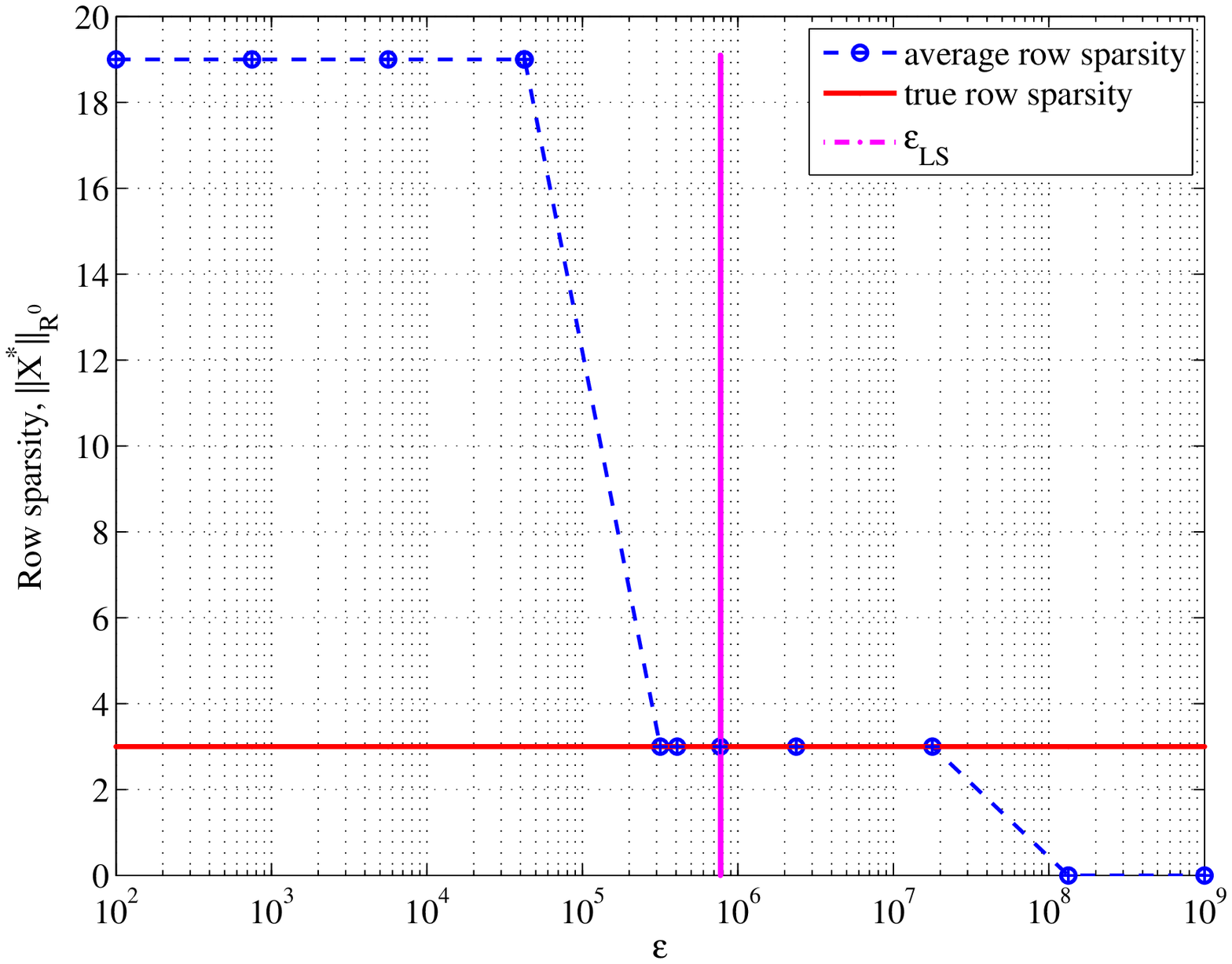}
  }
  \end{center}
  \caption{Sensitivity of the row sparsity recovery to the value of $\epsilon$.
We scan through a range of values of $\epsilon$ and plot the mean and the standard deviation of the recovered row sparsity for LSCC-L12M problem. For Fig.~\ref{fig6a} we used intensity parameter $\theta_1 = 10^2$, for Fig.~\ref{fig6b} we used intensity parameter $\theta_2 = 10^3$.}
  \label{fig6}
\end{figure}

Theorem \ref{thmLS} suggests that by selecting $\epsilon_{LS}$ according to Proposition \ref{propRadiusLS}, row sparsity minimization guarantees exact row sparsity and row sparsity pattern recovery with probability at least $1-p$. Here, we set $p = 0.05.$ To examine the effect of varying corresponding $\epsilon$ on row sparsity and row sparsity pattern recovery accuracy, we consider the following setup. We define a range of values for $\epsilon$, $[10^2,10^9]$ which includes the value of $\epsilon_{LS}$ from Proposition \ref{propRadiusLS}. We use synthetic data generated as described in Section $\ref{ssec:datageneration}$. To incorporate the effect of noise, we define two different intensity parameters $\theta_1 = 10^2$ and $\theta_2 = 10^3$. For each value of $\epsilon$ in the range we solve the LSCC-L12M problem using the method described in Section \ref{sec:gradappr}. We use $30$ iterations for the binary search in Algorithm \ref{algOuterIter} and $3000$ iterations in Algorithm \ref{backtracking}. Then we evaluate the row sparsity and row sparsity pattern of the corresponding recovered matrix $X^{LS}$. The row sparsity evaluation is done by counting the number of non-zero rows of the solution matrix. We threshold the solution matrix to avoid miscounting due to numerical errors. The threshold parameter is defined as $1\%$ of the $\gamma(\tilde X).$ To evaluate the row sparsity pattern we compare row sparsity pattern of the solution matrix with the row sparsity pattern of the true matrix $\tilde X$. We repeat this experiment ten times, regenerating a new sample measurement matrix $Y$ each time. Using ten runs for each value of $\epsilon$, we record the mean and the standard deviation of the solution row sparsity obtained by solving LSCC-L12M. Figure \ref{fig6} shows row sparsity mean and standard deviation as a function of $\epsilon$. Figure \ref{fig6a} corresponds to the lower SNR environment with $\theta = 10^2$ and Figure \ref{fig6b} corresponds to the higher SNR environment with $\theta = 10^3$. On both plots we present two additional lines. The horizontal line indicates the true row sparsity of the initial matrix $\tilde X$. The vertical line indicates value of $\epsilon=\epsilon_{LS}$ found by the Proposition \ref{propRadiusLS}. Figure \ref{fig6} supports Theorem \ref{thmLS} by indicating that the choice of $\epsilon = \epsilon_{LS}$ leads to exact row sparsity pattern recovery. Since $\epsilon$ depends on $Y$, its value varies from one run to another. Consequently, ten nearly identical vertical lines are plotted in Figures \ref{fig6a} and \ref{fig6b}.

Notice also that there is only a small range of $\epsilon$ where the row sparsity and the row sparsity pattern can be recovered correctly and as $\epsilon$ deviates from the $\epsilon_{LS}$, the true row sparsity of matrix $X$ can no longer be recovered. Intuitively, when we increase $\epsilon$ the confidence-constrained set may include matrices with row sparsity lower than sparsity that are not in $\gamma$ - neighborhood of matrix $\tilde X$. Hence, the row sparsity minimization inside such confidence set may lead to recovery of a matrix with lower row sparsity. On the other hand, as we decrease $\epsilon$, the confidence set may not include the true matrix $\tilde X$, therefore, the row sparsity of the recovered matrix may be higher than the row sparsity of matrix $\tilde X$. We observe that the choice of $\epsilon = \epsilon_{LS}$ given by Proposition \ref{propRadiusLS} can be used as a reasonable rule of thumb for solving the LSCC-L12M problem in the convex setting. Notice also that the solution's sensitivity to the intensity value $\theta$ is affected by SNR: for a lower SNR scenario with $\theta = 10^2$, the range of $\epsilon$ for which the row sparsity pattern can be recovered correctly is smaller than the corresponding range for a higher SNR scenario with $\theta = 10^3$.

\begin{figure}
  \begin{center}
  \subfigure[]{
  \label{fig7a}
  \includegraphics[scale=0.44]{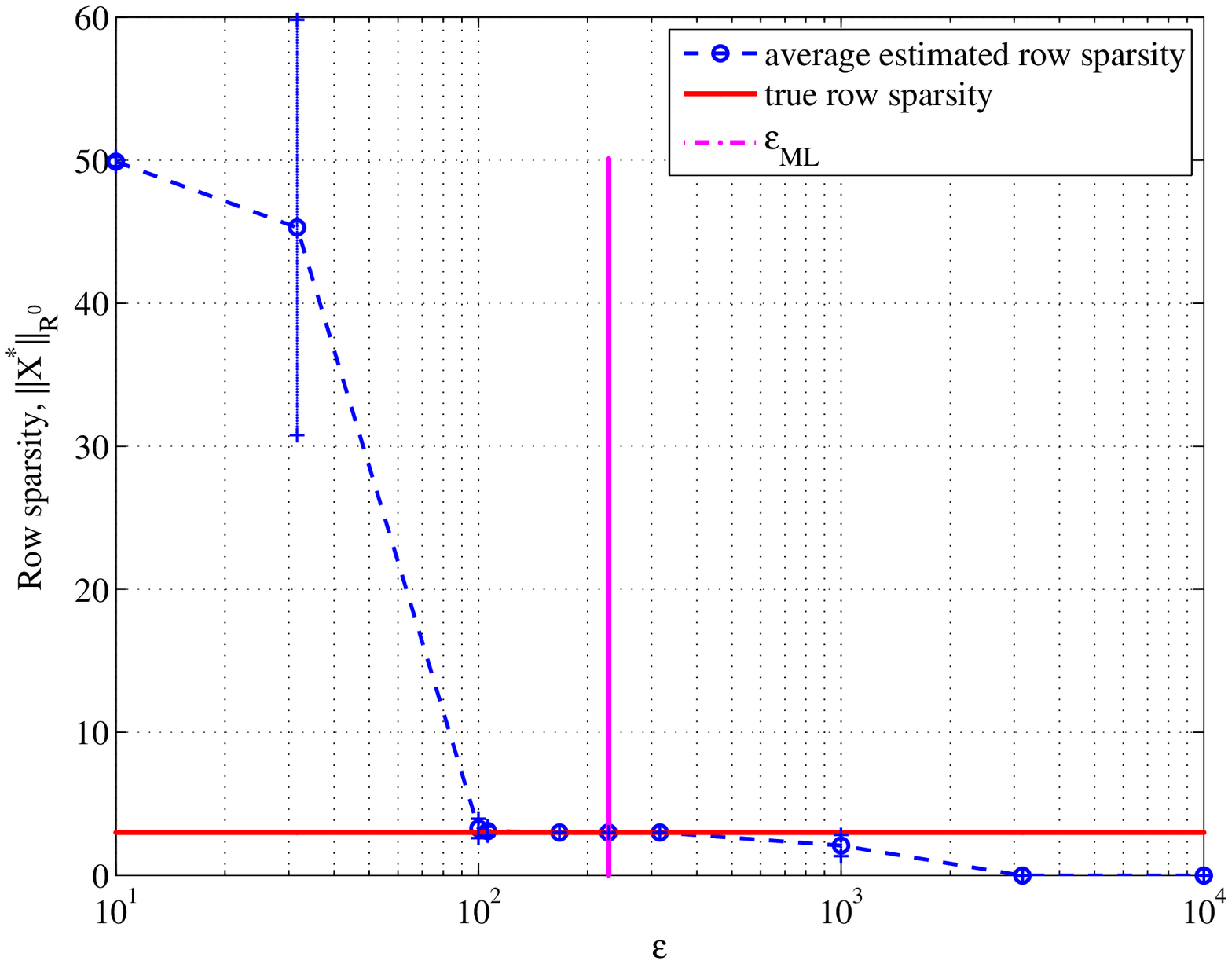}
  }
  \subfigure[]{
  \label{fig7b}
  \includegraphics[scale=0.44]{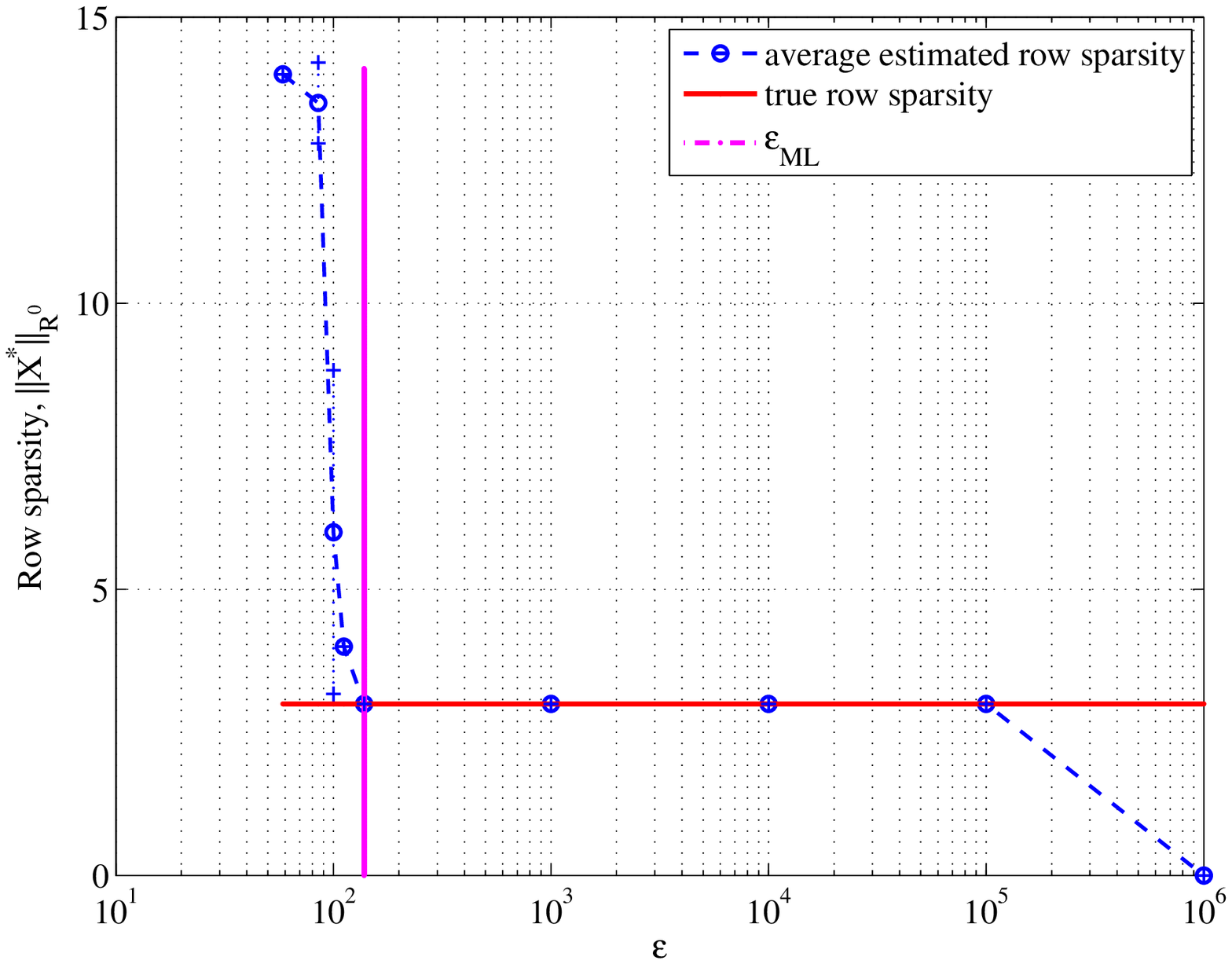}
  }
  \end{center}
  \caption{Sensitivity of the row sparsity recovery to the value of $\epsilon$. We scan through a range of values of $\epsilon$ and plot the mean and the standard deviation of the recovered row sparsity for MLCC-L12M problem. For Fig.~\ref{fig7a} we used intensity parameter $\theta_1 = 10^1$, for Fig.~\ref{fig7b} we used intensity parameter $\theta_2 = 10^3$.}
  \label{fig7}
\end{figure}

Similar discussion applies to Figure \ref{fig7}, where we present the row sparsity mean and the standard deviation as a function of $\epsilon$ for the MLCC-L12M problem. Figure \ref{fig7a} corresponds to a lower SNR environment with $\theta = 10^1$ and Figure \ref{fig7b} corresponds to a higher SNR environment with $\theta = 10^3$. The horizontal line indicates the true row sparsity of initial matrix $\tilde X$. Notice that $\epsilon_{ML}$ found by Proposition \ref{propRadiusML} is independent of observation matrix $Y$. Therefore, in both figures there is only one vertical line that depicts the value of $\epsilon_{ML}$. Figure \ref{fig7} supports Theorem \ref{thmML} by indicating that the choice of $\epsilon = \epsilon_{ML}$ leads to exact row sparsity pattern recovery.

\subsection{Probability of correct row sparsity pattern recovery}

In this section, we investigate the probability of correct row sparsity pattern recovery, i.e., when the quantity and the location of the non-zero rows are found correctly. For both LSCC-L12M (\ref{CCLScvx}) and MLCC-L12M (\ref{CCMLcvx}) problems, we compare the empirical probability of correct row sparsity pattern recovery with the sufficient conditions proposed by Theorem \ref{thmLS} and Theorem \ref{thmML}, respectively. Note that for both Theorem \ref{thmLS} and Theorem \ref{thmML} we set $p = 0.05$. Our objective is to show that the exact row sparsity and row sparsity pattern recovery conditions proposed in Theorems \ref{thmLS} and \ref{thmML} are sufficient in the case when we replace row sparsity minimization with $\ell_{1,2}$-norm minimization.

We use synthetic data generated according to the description in the Section $\ref{ssec:datageneration}$. For both LSCC-L12M (\ref{CCLScvx}) and MLCC-L12M (\ref{CCMLcvx}) problems we define the range of the intensity values $\theta$, that control the strength of the signal, by generating twenty logarithmically spaced points in the interval $[10^{-2},10^9]$. Iteratively, we scan through the range of intensity values $\theta$, and for each $\theta$ we change the strength of the signal $\tilde X \gets \theta \tilde X$, and regenerate new matrix of Poisson counts $Y$. Then, we run the corresponding recovery algorithm with $30$ iterations of the binary search in Algorithm \ref{algOuterIter} and $3000$ iterations of Algorithm \ref{backtracking}. For each intensity value $\theta$, we estimate corresponding solution matrix $X^*$. We repeat this experiment ten times for each framework. For each value of $\theta$ we calculate the number of times the row sparsity pattern was found correctly. Here we use an $\zeta$-row sparsity pattern measure, i.e., we consider a row $j$ of matrix $X^*$ to be non-zero, if its Euclidian norm is greater than $\zeta$. In other words, $\mathrm{RSupp}_{\zeta}(X) = \{j : \|(x^{j})^{\mathrm{T}}\|_2 > \zeta]\}$. Averaging over ten runs, we obtain the empirical probability of correct row sparsity pattern recovery.

\begin{figure}
  \begin{center}
  \subfigure[LSCC-L12M]{
  \label{fig4a}
  \includegraphics[scale=0.44]{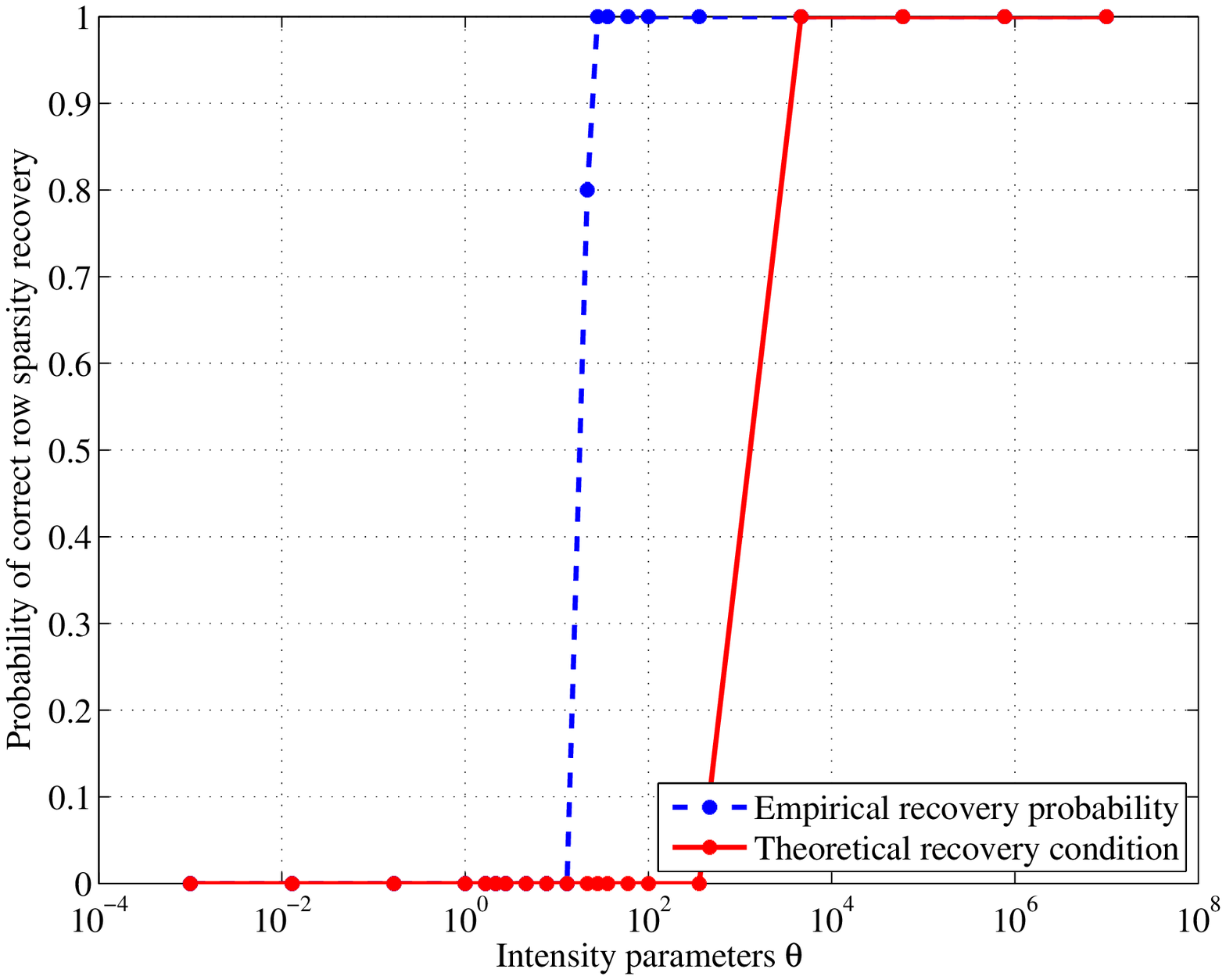}
  }
  \subfigure[MLCC-L12M]{
  \label{fig4b}
  \includegraphics[scale=0.44]{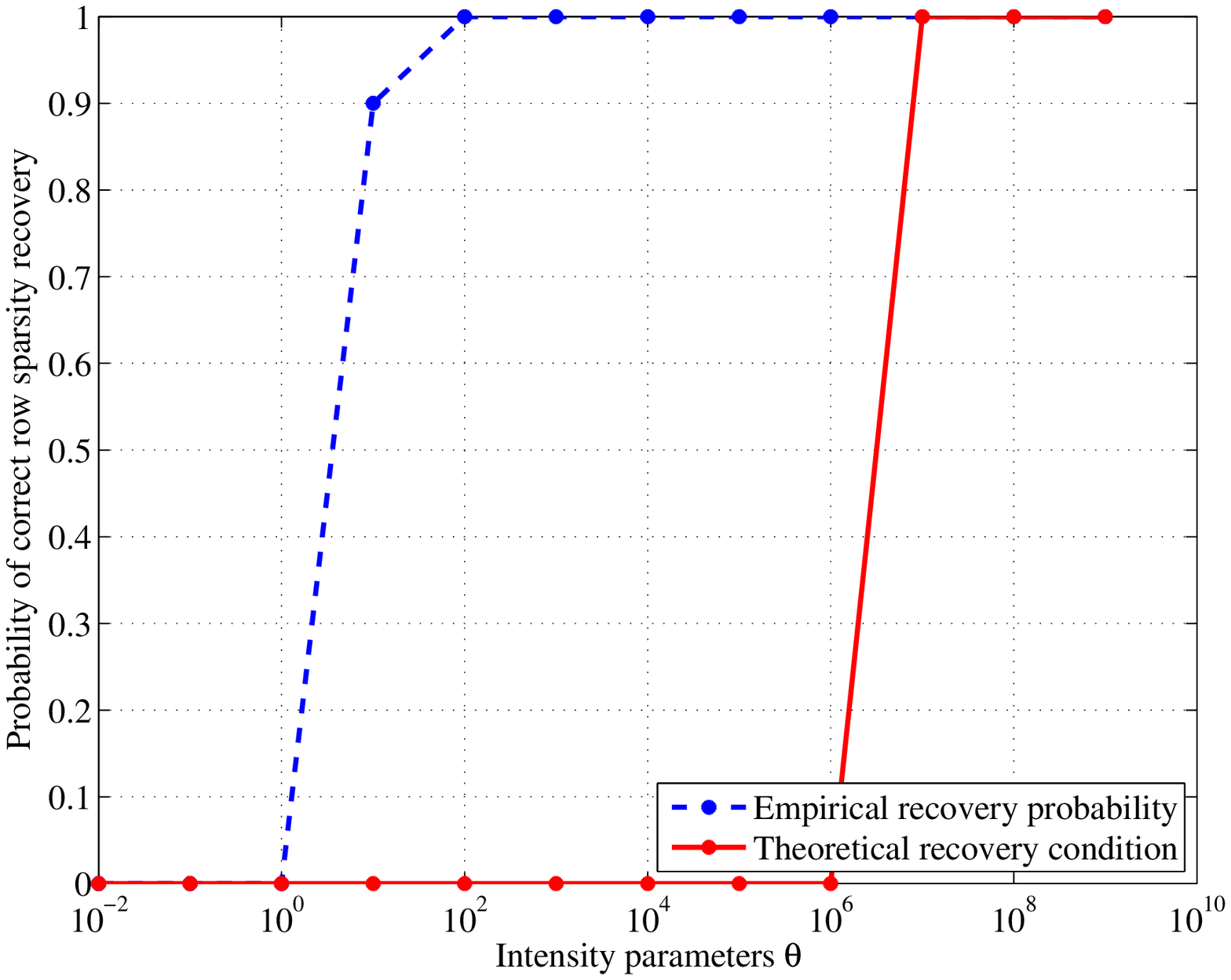}
  }
  \end{center}
  \caption{Empirical probability of row sparsity pattern recovery against the intensity values $\theta$ for problems LSCC-L12M \ref{fig4a} and MLCC-L12M \ref{fig4b}. For every value of $\theta$, we plot the theoretical recovery condition provided by Theorems \ref{thmLS} and \ref{thmML}, respectively.}
  \label{fig4}
\end{figure}

In Figures \ref{fig4a} and \ref{fig4b}, we depict the empirical probability of row sparsity pattern recovery against intensity values $\theta$ for both LSCC-L12M (\ref{CCLScvx}) and MLCC-L12M (\ref{CCMLcvx}) problems, respectively. On both figures the dashed lines correspond to the empirical estimate of probability of correct row sparsity pattern recovery. Solid lines represent theoretical row sparsity pattern recovery conditions provided by corresponding Theorems \ref{thmLS} and \ref{thmML}, respectively.

For both LS and ML frameworks, the area for exact row sparsity recovery probability covers the success region, the sufficient conditions proposed by corresponding Theorems \ref{thmLS} and \ref{thmML} appear to hold for the heuristic replacement of $\|\cdot\|_{R^0}$ norm minimization with $\|\cdot\|_{1,2}$ norm minimization.

Note that the gap in Figure \ref{fig4b} is larger. Although, for both LS and ML frameworks we used similar bounding technique for deriving conditions for row sparsity and row sparsity pattern recovery in Theorems \ref{thmLS} and \ref{thmML}, the derivation of the sufficient conditions for ML framework involved larger number of bounds. This gap suggests that there is a way to improve the bound.

\section{Conclusion}
\label{sec:conclusion}

In this paper, we introduces the framework of confidence-constrained row sparsity minimization to recover the true row sparsity pattern under the Poisson noise model. We formulated the problem in a tuning-free fashion, such that the objective function controls the row sparsity and constraints control the data fit. Using a statistical analysis of the Poisson noise model in both the LS and ML frameworks, we determined the value for the constrained parameters $\epsilon_{LS}$ and $\epsilon_{ML}$. Moreover, we derive the conditions under which the exact row sparsity and row sparsity pattern can be recovered. The proposed formulas for $\epsilon_{LS}$ and $\epsilon_{ML}$ are shown to be effective in selecting the values of the tuning parameters to yield the correct row sparsity pattern recovery for the convex relaxation of the problem.

This paper motivates the concept of statistically provable tuning-free approach for row sparsity pattern recovery in noisy conditions. An important extension to this work could be to identify the conditions for which both $\|\cdot\|_{R^0}$ norm and $\|\cdot\|_{1,2}$ norm minimization provide the same results.

\section{Acknowledgment}

This work is partially supported by the National Science Foundation grants CCF-1254218 and CNS-0845476.

\section{Appendix }
\label{sec:APPA}

\subsection{Proof of Proposition \ref{propUpperBound}}

\proof

Let $X^*$ be the minimizer of the $\|\cdot\|_{R^0}$-norm over the set $S$, i.e., $X^* \in \arg \min_{X\in S}\|X\|_{R^0}$. Therefore, $\|X^*\|_{R^0} \le \|X\|_{R^0}$ for all $X \in S$. Specifically, since $\tilde X \in S$, then $\|X^*\|_{R^0} \le \|\tilde X\|_{R^0}$.
\endproof

\subsection{Proof of Proposition \ref{propRadiusLS}}
\proof
Let $\Lambda = \sum_{i=1}^{N} \sum_{j=1}^{M} \lambda_{i}(j)$. Consider an auxiliary random variable $\psi = \sum_{i=1}^{N}\sum_{j=1}^{M} y_{i}(j),$ where $y_i(j)\sim \mathrm{Poisson}(\lambda_{i}(j))$, $A_{i} \tilde x_{i} = \lambda_{i}$, $\forall i=\overline{1,N},j=\overline{1,M}$. Since measurements $y_{i}(j)$ are independent $\forall i=\overline{1,N},j=\overline{1,M}$, therefore $\psi$ is a Poisson random variable with parameter $\Lambda$, i.e., $\psi \sim \mathrm{Poisson}(\Lambda).$ Moreover, $\mathbb{E}[\psi] = \mathbb{VAR}[\psi] = \Lambda.$

Now consider a random variable $S$ defined as follows: $S = \sum_{i=1}^{N} \|A_{i} \tilde x_{i}-y_{i}\|_{2}^{2}.$
To find the statistics of random variable $S$, we first recall that for a Poisson random variable $y_{i}(j)$ with parameter $\lambda_{i}(j)$, $\forall i=\overline{1,N},j=\overline{1,M}$, has the following moments: $\mathbb{E}[y_{i}(j)] = \lambda_{i}(j)$, $\mathbb{E}[y_{i}^{2}(j)] = \lambda_{i}^{2}(j)+\lambda_{i}(j)$, $\mathbb{E}[y_{i}^{3}(j)] = \lambda_{i}^{3}(j)+3\lambda_{i}^{2}(j)+\lambda_{i}(j)$, $\mathbb{E}[y_{i}^{4}(j)] = \lambda_{i}^{4}(j)+6\lambda_{i}^{3}(j)+7\lambda_{i}^{2}(j)+\lambda_{i}(j).$
Therefore, the expectation  and the variance of random variable $S$ can be found as follows:
$$\mathbb{E}[S] = \mathbb{E}\Big[\sum_{i=1}^{N} \|A_{i}\tilde x_{i}-y_{i}\|_{2}^{2}\Big] = \sum_{i=1}^{N} \sum_{j=1}^{M}\mathbb{E}[(\lambda_{i}(j)-y_{i}(j))^2] = \sum_{i=1}^{N} \sum_{j=1}^{M}\mathbb{VAR}[y_{i}(j)] = \sum_{i=1}^{N} \sum_{j=1}^{M} \lambda_{i}(j) = \Lambda.$$
$$\mathbb{VAR}[S] 
= \sum_{i=1}^{N} \sum_{j=1}^{M}\mathbb{E}\left[ \left( (\lambda_{i}(j)-y_{i}(j))^2-\lambda_{i}(j)\right)^2\right] = \sum_{i=1}^{N} \sum_{j=1}^{M}\left(2\lambda_{i}(j)^2+\lambda_{i}(j)\right).$$

Now, applying one-sided Chebyshev's inequality we can get a probabilistic upper bound on the random variable $S$:
$\mathbb{P}\left(S \le \mathbb{E}[S]+k\sqrt{\mathbb{VAR}[S]} \right) \ge 1-\frac{1}{1+k^2}$
or
\begin{equation}
\label{cheb1}
\mathbb{P}\left(\sum_{i=1}^{N}\|A_{i} \tilde x_{i}-y_{i}\|_{2}^{2} \le \sum_{i=1}^{N} \sum_{j=1}^{M} \lambda_{i}(j)+k\sqrt{\sum_{i=1}^{N} \sum_{j=1}^{M}\left(2\lambda_{i}(j)^2+\lambda_{i}(j)\right)} \right) \ge 1-\frac{1}{1+k^2}.
\end{equation}

Although (\ref{cheb1}) provides an in-probability bound on term $\sum_{i=1}^{N}\|A_{i} \tilde x_{i}-y_{i}\|_{2}^{2}$, the bound depends on the unknown terms $\lambda_i(j)$, $\forall i = \overline{1,N}, j = \overline{1,M}$. Since our goal is to have a parameter-free bound, we proceed by bounding the term $\sum_{i=1}^{N} \sum_{j=1}^{M} \lambda_{i}(j)$ with a function that depends only on $\psi$ and is independent of $\lambda_i(j)$, $\forall i = \overline{1,N}, j = \overline{1,M}$.

In order to obtain such a bound, we first acquire inequality $\mathbb{P}\left(\psi \ge \Lambda-k\sqrt{\Lambda} \right) \ge 1-\frac{1}{1+k^2}$, using one-sided Chebyshev's inequality. Then we consider inequality
\begin{equation}
\label{hlpineq1}
\psi \ge \Lambda-k\sqrt{\Lambda}.
\end{equation}
Since $k\ge0$, and $\Lambda \ge 0$ we can square both sides and get an equivalent to (\ref{hlpineq1}) quadratic inequality:
$0\ge \Lambda^2-\Lambda\left(2\psi+k^2\right)+\psi^2.$
Solving this inequality provides two solutions for $\Lambda$: $\Lambda_{1} = \psi+\frac{k^2}{2}+\sqrt{k^2\psi+\frac{k^4}{4}}, \Lambda_{2} = \psi+\frac{k^2}{2}-\sqrt{k^2\psi+\frac{k^4}{4}}.$
Therefore, the solution for inequality (\ref{hlpineq1}) are all $\Lambda$, such that $\left(\Lambda \ge \Lambda_{2}\right)\cap\left(\Lambda \le \Lambda_{1}\right)$.
Hence,
$\mathbb{P}\left(\psi \ge \Lambda-k\sqrt{\Lambda} \right) = \mathbb{P}\left(\left(\Lambda \ge \Lambda_{2}\right)\cap\left(\Lambda \le \Lambda_{1}\right) \right).$
By properties of probabilities we obtain a following bound:
$\mathbb{P}\left(\Lambda \le \Lambda_{1}\right) \ge \mathbb{P}\left(\left(\Lambda \ge \Lambda_{2}\right)\cap\left(\Lambda \le \Lambda_{1}\right) \right),$
therefore,
$\mathbb{P}\left(\Lambda \le \Lambda_{1}\right) \ge \mathbb{P}\left(\left(\Lambda \ge \Lambda_{2}\right)\cap\left(\Lambda \le \Lambda_{1}\right) \right) = \mathbb{P}\left(\psi \ge \Lambda-k\sqrt{\Lambda} \right) \ge 1-\frac{1}{1+k^2},$
i.e.,
$\mathbb{P}\left(\Lambda \le \Lambda_{1}\right) \ge 1-\frac{1}{1+k^2}.$
Rewriting $\Lambda_{1}$ in terms of $\psi$ and $k$, we obtain
\begin{equation}
\label{hlpineq2}
\mathbb{P}\left(\Lambda \le \psi+\frac{k^2}{2}+\sqrt{k^2\psi+\frac{k^4}{4}}\right) \ge 1-\frac{1}{1+k^2}.
\end{equation}
Taking into the account that $\Lambda = \sum_{i=1}^{N}\sum_{j=1}^{M} \lambda_{i}(j)$ and $\psi = \sum_{i=1}^{N}\sum_{j=1}^{M} y_{i}(j)$,
we can rewrite (\ref{hlpineq2}) in terms of $y_{i}(j)$ and $\lambda_{i}(j)$ as follows:
\begin{equation}
\label{cheb2}
\mathbb{P}\left(\sum_{i=1}^{N}\sum_{j=1}^{M} \lambda_{i}(j) \le \sum_{i=1}^{N}\sum_{j=1}^{M} y_{i}(j)+\frac{k^2}{2}+\sqrt{k^{2}\sum_{i=1}^{N}\sum_{j=1}^{M} y_{i}(j)+\frac{k^4}{4}} \right) \ge 1-\frac{1}{1+k^2}.
\end{equation}
Notice, that inequality (\ref{cheb2}) provides an upper bound on unknown sum $\sum_{i=1}^{N}\sum_{j=1}^{M} \lambda_{i}(j)$ by a function of known variables $\sum_{i=1}^{N}\sum_{j=1}^{M} y_{i}(j)$ only.

Next we combine two inequalities (\ref{cheb1}) and (\ref{cheb2}) as follows.
First, for simplicity of notation let
$$F\left(\Lambda\right) = \sum_{i=1}^{N} \sum_{j=1}^{M} \lambda_{i}(j)+k\sqrt{\sum_{i=1}^{N} \sum_{j=1}^{M}\left(2\lambda_{i}(j)^2+\lambda_{i}(j)\right)},$$
$$G\left(\psi\right) = \sum_{i=1}^{N}\sum_{j=1}^{M} y_{i}(j)+\frac{k^2}{2}+\sqrt{k^{2}\sum_{i=1}^{N}\sum_{j=1}^{M} y_{i}(j)+\frac{k^4}{4}}.$$
To find $\mathbb{P}\left(S \le F\left(G\left(\psi\right)\right)\right)$ first recall that by properties of probabilities we have
$\mathbb{P}\left(S \le F\left(G\left(\psi\right)\right)\right) \ge \mathbb{P}\left(S \le F\left(\Lambda\right) \le F\left(G\left(\psi\right)\right)\right).$
Also,
$\mathbb{P}\left(\overline{S \le F\left(\Lambda\right) \le F\left(G\left(\psi\right)\right)}\right) = $ $\mathbb{P}(\left[S\ge F\left(\Lambda\right)\right]\cup [F\left(\Lambda\right)\ge $ $F\left(G\left(\psi\right)\right)])$
$\le \mathbb{P}\left(S\ge F\left(\Lambda \right)\right) + $ $ \mathbb{P}(F\left(\Lambda \right) \ge $ $F\left(G\left(\psi \right)\right)) =$ $\mathbb{P}\left(S\ge F\left(\Lambda \right)\right) +$ $ \mathbb{P}\left(\Lambda \ge G\left(\psi \right)\right) \le $ $ \frac{1}{1+k^2}+\frac{1}{1+k^2} = \frac{2}{1+k^2},$
therefore,
$\mathbb{P}\left(S \le F\left(G\left(\psi\right)\right)\right)\ge \mathbb{P}\left(S \le F\left(\Lambda\right) \le F\left(G\left(\psi\right)\right)\right) \ge 1-\frac{2}{1+k^2}.$
Now let $k = \sqrt{\frac{2}{p}-1}$, and set $\epsilon_{LS}=$
$$F\left(G\left(\psi \right)\right) = \psi+\frac{k^{2}}{2}+\sqrt{\psi k^2+\frac{k^{4}}{4}}+k \sqrt{2\psi^{2}+\psi(4k^{2}+1)+k^{4}+\frac{k^{2}}{2}+
\left(4\psi + 2k^2+1\right)\sqrt{\psi k^2+\frac{k^{4}}{4}}},$$
therefore obtaining
$\mathbb{P}\left(\sum_{i=1}^{N} \|A_i \tilde x_i-y_i\|_{2}^{2} \le \epsilon_{LS}\right) \ge 1-p.$
In other words, with probability at least $1-p$,
$ \tilde X \in S_{\epsilon_{LS}}.$
\endproof

\subsection{Proof of Proposition \ref{propRadiusML}}

\proof
Note that each term in the sum $\sum_{i=1}^{N} I(y_{i}||\lambda_{i})$ can be written as a sum of independent random variables of the type $I(y_{i}(j)||\lambda_{i}(j))$, i.e.:
$$\sum_{i=1}^{N} I(y_{i}||\lambda_{i}) = \sum_{i=1}^{N}\sum\limits_{j=1}^{M} I(y_{i}(j)||\lambda_{i}(j))= \sum_{i=1}^{N}\sum\limits_{j=1}^{M}\Big(y_{i}(j)\log(y_{i}(j))-y_{i}(j)\log(\lambda_{i}(j))-y_{i}(j)+\lambda_{i}(j)\Big).$$
For simplicity, we omit the dependence on $i$ and $j$ and focus on the term $I(y||\lambda)$.

The expectation of $I(y||\lambda)$ can be calculated as follows: $$\mathbb{E}\left[I(y||\lambda)\right]=\mathbb{E}\left[y\log(y)-y\log(\lambda)-y+\lambda\right] =-\lambda\log(\lambda)+\mathbb{E}\left[y\log(y)\right] $$
$$=-\lambda\log(\lambda)+\sum_{y=0}^{\infty}\frac{y\log(y)\exp(-\lambda)\lambda^{y}}{(y)!} =-\lambda\log(\lambda)+\lambda\exp(-\lambda)\sum_{y=0}^{\infty}\frac{\log(y+1)\lambda^{y}}{(y)!}.$$

Note, by the ratio test series $\sum_{y=0}^{\infty}\frac{\log(y+1)\lambda^{y}}{(y)!}$ converges absolutely:
$L = \lim_{n\rightarrow\infty}\left|\frac{a_{n+1}}{a_{n}}\right|= $  $\lim_{n\rightarrow\infty} $ $\left|\frac{\lambda\log(n+2)}{n\log(n+1)}\right| =0<1.$
Therefore, the expectation $\mathbb{E}\left[I(y||\lambda)\right]$ is finite.
Moreover, from the Taylor series expansion of $I(y||\lambda)$ given as:
\begin{equation}
\label{hlpeq9}
I(y||\lambda) = \sum_{k=2}^{\infty} \frac{(-1)^k (y-\lambda)^k}{(\lambda)^{k-1}k(k-1)},
\end{equation}
we conclude that $\mathbb{E}\left[I(y||\lambda)\right] = 0.5 + O(1/\lambda).$
Numerical evaluation shows that $\forall \lambda >0$ expectation $\mathbb{E}\left[I(y||\lambda)\right]$ can be bounded as follows:
\begin{equation}
\label{hlpineq3}
\mathbb{E}\left[I(y||\lambda)\right] < C_{\mu} \approx 0.5801.
\end{equation}

Consider the variance of the term $I(y||\lambda)$:
$$\mathbb{VAR}\bigl[I(y||\lambda)\bigr] = \mathbb{E}\Bigl[\bigl(I(y||\lambda)\bigr)^2\Bigr] -\Big(\mathbb{E}\left[I(y||\lambda)\right]\Big)^2
 = \mathbb{E}\Bigl[\bigl(y\log(y)-y\log(\lambda)-y+\lambda\bigr)^2\Bigr] -\Big(\mathbb{E}\left[I(y||\lambda)\right]\Big)^2 $$
$$=\lambda+2\lambda\log(\lambda)+\Big(\lambda+\lambda^{2}\Big)\log^{2}(\lambda) +  \mathbb{E}\left[y^{2}\log^{2}(y)\right]
 -2\Big(1+\log(\lambda)\Big)\mathbb{E}\left[y^{2}\log(y)\right] + 2\lambda\mathbb{E}\left[y\log(y)\right].$$
Variance $\mathbb{VAR}\left[I(y||\lambda)\right]$ is finite, due to the ratio test, all three terms $\mathbb{E}\left[y^{2}\log^{2}(y)\right]$, $\mathbb{E}\left[y^{2}\log(y)\right]$, and $\mathbb{E}\left[y^{2}\log(y)\right]$ are finite.
Moreover, from the Taylor series expansion of $I(y||\lambda)$ given by (\ref{hlpeq9}) and from the fact that
$\mathbb{E}\left[\left(I(y||\lambda)\right)^2\right] = 3/4 + O(1/\lambda)$,
we conclude that $\mathbb{VAR}\left[I(y||\lambda)\right] = 3/4 - (0.5)^2 + O(1/\lambda) = 0.5 + O(1/\lambda).$
Numerical evaluation shows that $\forall \lambda >0$ variance $\mathbb{VAR}\left[I(y||\lambda)\right]$ can be bounded as follows:
\begin{equation}
\label{hlpineq4}
\mathbb{VAR}\left[I(y||\lambda)\right]  < C_{\sigma^2} \approx 0.5178.
\end{equation}

\begin{figure}
  \begin{center}
  \includegraphics[scale=0.44]{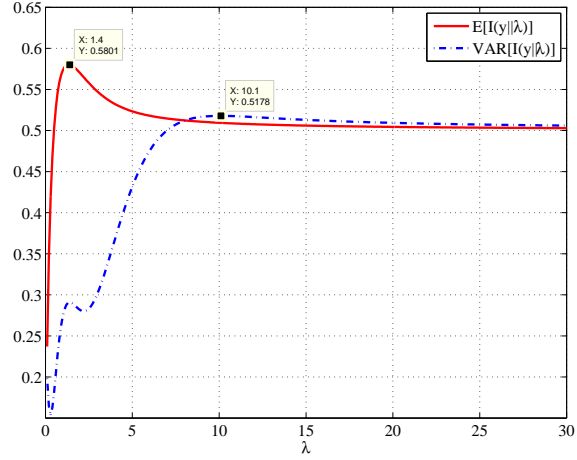}\\
  \end{center}
  \caption{The expectation (solid) and the variance (dash-dotted) of the I-divergence $I(y||\lambda)$ between a Poisson RV $y$ and its mean $\lambda$.}
  \label{evardij}
\end{figure}

Figure (\ref{evardij}) depicts the expectation and the variance of the term $I(y||\lambda)$, $\forall \lambda >0$.

Now we return to the notation that depends on the indexes $i$ and $j$. By one-sided Chebyshev's inequality we obtain the following inequality:
$$\mathbb{P}\left(\sum_{i=1}^{N}I(y_{i}||\lambda_{i}) \le \mathbb{E}\left[\sum_{i=1}^{N}I(y_{i}||\lambda_{i})\right]+k\sqrt{\mathbb{VAR}\left[\sum_{i=1}^{N}I(y_{i}||\lambda_{i})\right]}\right)\ge 1-\frac{1}{1+k^2}.$$
From the independence of terms $I(y_{i}(j)||\lambda_{i}(j))$, it follows that:
$$\mathbb{E}\left[\sum_{i=1}^{N}I(y_{i}||\lambda_{i})\right]=\sum\limits_{i=1}^{N} \sum\limits_{j=1}^{M} \mathbb{E}\left[I(y_{i}(j)||\lambda_{i}(j))\right], \mathbb{VAR}\left[\sum_{i=1}^{N}I(y_{i}||\lambda_{i})\right]=\sum\limits_{i=1}^{N}\sum\limits_{j=1}^{M} \mathbb{VAR}\left[I(y_{i}(j)||\lambda_{i}(j))\right].$$
Therefore, taking into account numerical bounds (\ref{hlpineq3}), (\ref{hlpineq4}) we obtain:
$$\mathbb{E}\left[\sum_{i=1}^{N}I(y_{i}||\lambda_{i})\right]+k\sqrt{\mathbb{VAR}\left[\sum_{i=1}^{N}I(y_{i}||\lambda_{i})\right]} $$
$$\le \sum_{i=1}^{N}\sum_{j=1}^{M}\mathbb{E}\left[I(y_{i}(j)||\lambda_{i}(j))\right] + k\sqrt{\sum_{i=1}^{N}\sum_{j=1}^{M}\mathbb{VAR}\left[I(y_{i}(j)||\lambda_{i}(j))\right]} \le C_{\mu}MN+k\sqrt{C_{\sigma^2}MN}.$$
Now let $k = \sqrt{\frac{1}{p}-1}$ and set $\epsilon_{ML} = C_{\mu}MN+k\sqrt{C_{\sigma^2}MN}$.
Hence,
$\mathbb{P}\left(\sum_{i=1}^{N}I(y_{i}||\lambda_{i}) \le \epsilon_{ML} \right)\ge 1-p.$
In other words, with probability at least $1-p$, $ \tilde X \in S_{\epsilon_{ML}}. $
\endproof

\subsection{Proof of Proposition \ref{propLowerBound}}

\proof (By contradiction) Assume that there exist a matrix $X^*$ such that $\|X^*\|_{R^0} < \|\tilde X\|_{R^0}$ and $\|X^*-\tilde X\|_F < \gamma(\tilde X)$. Then, there exist a row in $\tilde X$ that is non-zero while the corresponding row in $X^*$ is zero. Without loss of generality, we assume that this is the $l'$th row of $\tilde X$. Hence,
$$\| X^* - \tilde X \|_F \ge \min_{X',\| X'\|_{R^0} < \|\tilde X\|_{R^0}}\| X' - \tilde X \|_F = \min_{X',\| X'\|_{R^0} < \|\tilde X\|_{R^0}}\sqrt{ \sum_{l=1}^{K}\|e_l^{\mathrm{T}}\left(X' - \tilde X\right)\|_2^2 } \ge$$
$$\min_{X',\| X'\|_{R^0} < \|\tilde X\|_{R^0}} \|e_{l'}^{\mathrm{T}}\left(X' - \tilde X\right)\|_2 = \|e_{l'}^{\mathrm{T}} \tilde X\|_2 \ge \min_{l,\| e_l^{\mathrm{T}} \tilde X \|_2\neq 0} \| e_l^{\mathrm{T}} \tilde X \|_2 = \gamma(\tilde X).$$
But this contradicts the assumption $\|X^*-\tilde X\|_F < \gamma(\tilde X)$. Hence, the assumption $\|X^*\|_{R^0} < \|\tilde X\|_{R^0}$ is invalid. Therefore, $\|X^*\|_{R^0} \ge \|\tilde X\|_{R^0}$.
\endproof

\subsection{Proof of Theorem \ref{thmLS}}

\proof
{\em Exact row sparsity.}

Let matrix $X^{LS}$ be a solution to LSCC-RSM optimization problem (\ref{CCLS}). Note that $\epsilon_{LS}$ is chosen such that $\mathbb{P}\left(\tilde X \in S_{\epsilon_{LS}}\right)\ge 1-p$. Therefore, by Proposition \ref{propUpperBound} $\|X^{LS}\|_{R^0} \le \|\tilde X\|_{R^0}$ with probability at least $1-p$.

Now, since $X^{LS}$ is the solution to LSCC-RSM optimization problem (\ref{CCLS}) then $X^{LS}\in S_{\epsilon_{LS}}$. Hence, the Frobenius norm of the difference between the solution matrix $X^{LS}$ and the true row sparse matrix $\tilde X$ can be bounded as follows:
$$ \| X^{LS} - \tilde X \|_F  =  \sqrt{\sum_{i=1}^{N} \| x^{LS}_{i}- \tilde x_{i} \|_2^2} \stackrel{\circled{1}}{\le}   \frac{\sqrt{ \sum_{i=1}^{N} \|A_i( x^{LS}_{i}- \tilde x_{i} )\|_2^2} }{1-\delta_{2s}}$$
$$ \stackrel{\circled{2}}{\le}  \frac{\sqrt{ \sum_{i=1}^{N}(\|A_i x^{LS}_{i} - y_i \|_2 + \|A_i \tilde x_{i} -y_i\|_2)^2}}{1-\delta_{2s}}  \stackrel{\circled{3}}{\le}  \frac{\sqrt{ 4 \epsilon_{LS}}}{ 1-\delta_{2s}}  <  \gamma_{LS}(\tilde X),$$
where $\circled{1}$ is due to fact that $A_i$ satisfies the $2s$-restricted isometry property $\forall i = \overline{1,N}$ and that sparsity of vector $x^{LS}_{i}- \tilde x_{i}$ is at most $2s$, $\forall i=\overline{1,N}$, i.e., $\|x^{LS}_{i}- \tilde x_{i}\|_0\le 2s$, $\circled{2}$ is due to the triangle inequality, and $\circled{3}$ is due to the definition of the constraints for LSCC-RSM optimization problem (\ref{CCLS}). Therefore, it follows that since $X^{LS} \in S_{\epsilon_{LS}}$ then $X^{LS}$ is in the $\gamma_{LS}(\tilde X)$-Frobenius ball neighborhood of $\tilde X$, i.e., $X^{LS} \in \mathrm{B}_{\gamma_{LS}(\tilde X)}^{F}(\tilde X)$. Hence, $S_{\epsilon_{LS}} \subseteq \mathrm{B}_{\gamma_{LS}(\tilde X)}^{F}(\tilde X).$ Therefore, by Proposition \ref{propLowerBound}: $\|X^{LS}\|_{R^0}\ge \|\tilde X\|_{R^0}$. Combining inequalities $\|X^{LS}\|_{R^0} \le \|\tilde X\|_{R^0}$ and $\|X^{LS}\|_{R^0}\ge \|\tilde X\|_{R^0}$ together, we conclude that $\|X^{LS}\|_{R^0} = \|\tilde X\|_{R^0}$ with probability at least $1-p$.

{\em Exact row sparsity pattern.}

Now, suppose that row sparsity pattern of the solution matrix $X^{LS}$ is different from the row sparsity pattern of $\tilde X$, i.e., $\mathrm{RSupp}(X^{LS}) \ne \mathrm{RSupp}(\tilde X),$ even if row sparsity is the same, $\|X^{LS}\|_{R^0} = \|\tilde X\|_{R^0}.$  Without loss of generality, we can assume that row-support of $X^{LS}$ and $\tilde X$ are different only by one row index:
$\mathrm{RSupp}(X^{LS}) \setminus \mathrm{RSupp}(\tilde X) = \{l_1,l_2\}$, i.e., $l_1$ row of $X^{LS}$ is zero, but $l_1$ row of $\tilde X$ is not zero and $l_2$ row of $X^{LS}$ is not zero, but $l_2$ row of $\tilde X$ is zero.
Now consider Frobenius norm of the difference between the solution matrix $X^{LS}$ and the true row sparse matrix $\tilde X$: $\| X^{LS} - \tilde X \|_F  = \sqrt{\sum_{l=1}^{K} \sum_{i=1}^{N} |X^{LS}(l,i)- \tilde X(l,i) |^2} = \sqrt{\sum_{l=1}^{K} \| ((x^{LS})^l)^{T}- (\tilde x^l)^{T} \|_2^2}.$
From the definition of $\gamma_{LS}(\tilde X)$, it follows that inequality $\| ((x^{LS})^{l_1})^{T}- (\tilde x^{l_1})^{T} \|_2^2 = \| (\tilde x^{l_1})^{T} \|_2^2 \ge (\gamma_{LS}(\tilde X))^2$ holds for $l_1$ term of the sum $\sum_{l=1}^{K} \| ((x^{LS})^l)^{T}- (\tilde x^l)^{T} \|_2^2$. Therefore,
$\| X^{LS}- \tilde X \|_F = \sqrt{\sum_{l=1}^{K} \| ((x^{LS})^l)^{T}- (\tilde x^l)^{T} \|_2^2} \ge \sqrt{\| ((x^{LS})^{l_1})^{T}- (\tilde x^{l_1})^{T} \|_2^2} \ge \sqrt{(\gamma_{LS}(\tilde X))^2} = \gamma_{LS}(\tilde X).$
This contradicts the assumption of the Proposition \ref{propLowerBound} that $\| X^{LS}- \tilde X \|_F<\gamma_{LS}(\tilde X)$ and hence contradicts the assumption that $\|X^{LS}\|_{R^0} = \|\tilde X\|_{R^0}.$ Therefore, $\mathrm{RSupp}(X^{LS}) = \mathrm{RSupp}(\tilde X).$
\endproof

\subsection{Proof of Theorem \ref{thmML}}

\subsubsection{Auxiliary lemmas}

Before we prove Theorem \ref{thmML}, we present several auxiliary lemmas. Lemmas \ref{Lemma1}-\ref{Lemma4} provide useful bounds on I-divergence and $\|\cdot\|_1$-norm terms.

\begin{lem}
\label{Lemma1}

Let $y,\lambda \in \mathbb{R}^{M\times 1}$ be such that $y(j), \lambda(j) \ge 0$, $\forall j = \overline{1,M}$. Let also, $Y=\|y\|_1$ and $\Lambda=\|\lambda\|_1$. Let $\epsilon >0$ be given. If $I(y||\lambda) \le \epsilon$ then the following inequalities hold:

\noindent\begin{tabularx}{\textwidth}{@{}XXX@{}}
\begin{equation}
\label{Idivbound1}
I\left(\frac{y}{Y}\Big|\Big|\frac{\lambda}{\Lambda}\right) \le \frac{\epsilon}{Y},
\end{equation} &
\begin{equation}
\label{Idivbound2}
I(Y||\Lambda) \le \epsilon.
\end{equation}
\end{tabularx}
\end{lem}

\proof
Notice that the I-divergence between $y$ and $\lambda$ can be expressed as
\begin{eqnarray}
\label{Idiv}
I(y||\lambda) = Y I\left(\frac{y}{Y}\Big|\Big|\frac{\lambda}{\Lambda} \right) + I(Y||\Lambda),
\end{eqnarray}
since
$$\sum_{j=1}^M \left(y(j) \log \left(\frac{y(j)}{\lambda(j)}\right) + \lambda(j)-y(j)\right)=
\sum_{j=1}^M \left(y(j) \log \left(\frac{y(j)/Y}{\lambda(j)/\Lambda}\right) +y(j) \log \left(\frac{Y}{\Lambda}\right)\right)  + \Lambda -Y$$
$$ = Y \sum_{j=1}^M \left(\frac{y(j)}{Y}\right) \log \left( \frac{y(j)/Y}{\lambda(j)/\Lambda}\right) +Y \log \left( \frac{Y}{\Lambda}\right)  + \Lambda -Y = Y I\left(\frac{y}{Y}\Big|\Big|\frac{\lambda}{\Lambda}\right) + I(Y||\Lambda).$$
Then two bounds (\ref{Idivbound1}) and (\ref{Idivbound2}) are obtained by bounding each nonnegative terms on the RHS of (\ref{Idiv}) by $\epsilon$.
\endproof

\begin{lem}
\label{Lemma2}

Let $y,\lambda \in \mathbb{R}^{M\times 1}$ be such that $y(j), \lambda(j) \ge 0$, $\forall j = \overline{1,M}$. Let also, $Y=\|y\|_1$ and $\Lambda=\|\lambda\|_1$. Let $\epsilon >0$ be given. If $I(y||\lambda) \le \epsilon$ the following inequality holds:
\begin{equation}
\label{l1normalizedbound}
\Big\| \frac{y}{Y} -  \frac{\lambda}{\Lambda} \Big\|_1 \le \sqrt{\frac{2 \epsilon}{Y}}.
\end{equation}
\end{lem}

\proof
First, notice that by Pinsker's inequality we have
$\Big\| \frac{y}{Y} -  \frac{\lambda}{\Lambda} \Big\|_1 \le \sqrt{2D_{KL}\left(\frac{y}{Y}\Big|\Big|\frac{\lambda}{\Lambda}\right)}. $
Second, since $\Big\|\frac{y}{Y}\Big\|_1 = \Big\|\frac{\lambda}{\Lambda}\Big\|_1 = 1$, then Kullback-Leibler divergence and I-divergence for vectors $\frac{y}{Y}$, $\frac{\lambda}{\Lambda}$ are equal, i.e., $D_{KL}\left(\frac{y}{Y}\Big|\Big|\frac{\lambda}{\Lambda}\right)=I\left(\frac{y}{Y}\Big|\Big|\frac{\lambda}{\Lambda}\right).$
Therefore, $\Big\| \frac{y}{Y} -  \frac{\lambda}{\Lambda} \Big\|_1 \le \sqrt{2D_{KL}\left(\frac{y}{Y}\Big|\Big|\frac{\lambda}{\Lambda}\right)} = \sqrt{2I\left(\frac{y}{Y}\Big|\Big|\frac{\lambda}{\Lambda}\right)} \le  \sqrt{\frac{2 \epsilon}{Y}}.$
The last inequality is obtained using the result (\ref{Idivbound1}) of Lemma (\ref{Lemma1}).
\endproof

\begin{lem}
\label{Lemma3}

Let $y,\lambda \in \mathbb{R}^{M\times 1}$ be such that $y(j), \lambda(j) \ge 0$, $\forall j = \overline{1,M}$. Let also, $Y=\|y\|_1$ and $\Lambda=\|\lambda\|_1$. Let $\epsilon >0$ be given, such that $I(y||\lambda) \le \epsilon.$ Then the following inequality is true:
\begin{equation}
\label{l1YLambda}
|Y-\Lambda| \le \sqrt{2\epsilon Y}+Y G\left(\frac{2\epsilon}{Y}\right),
\end{equation}
where auxiliary function $G$ is define as $G(z) = \frac{(1+\sqrt{2z})(z+\log(1+\sqrt{2z})-\sqrt{2z})}{\sqrt{2z}}.$
\end{lem}

\proof
To obtain bound (\ref{l1YLambda}), we consider two cases: when $\Lambda \ge Y$ and when $\Lambda < Y$. Then we bound quantity $|Y-\Lambda|$ in each case and combine two bounds together.

$\mathit{Case}$ $\mathit{1:}$ $\Lambda \ge Y$.
First notice that by Lemma (\ref{Lemma1}) inequality $I(Y||\Lambda) \le \epsilon$ holds and can be explicitly written as:
\begin{eqnarray}
\label{L3_1}
I(Y||\Lambda) = Y\log(Y/\Lambda)+\Lambda-Y \le \epsilon.
\end{eqnarray}
We reorganize (\ref{L3_1}) as follows:
\begin{eqnarray}
\label{L3_3}
-\log(1+ (\Lambda-Y)/Y)+(\Lambda-Y)/Y \le \epsilon/Y.
\end{eqnarray}
For simplicity of notation, let $t = (\Lambda-Y)/Y$ and $z = \epsilon/Y$. Notice that $t \ge 0$ because of the assumption $\Lambda \ge Y$. Then we can rewrite (\ref{L3_3}) as follows:
\begin{eqnarray}
\label{L3_2}
t \le z + \log(1+ t).
\end{eqnarray}
We propose the following bound $t \le \sqrt{2z} + G(z)$, where $G(z) = \frac{(1+\sqrt{2z})  (z + \log(1+\sqrt{2z}) -\sqrt{2z})}
{\sqrt{2z}} $. We obtain this bound as follows:
$$ t \le z + \log(1+\sqrt{2z}+ (t-\sqrt{2z})) =  z + \log(1+\sqrt{2z}) +\log \left(1 + \frac{t-\sqrt{2z}}{1+\sqrt{2z}}\right) \le  z + \log(1+\sqrt{2z}) + \frac{t-\sqrt{2z}}{1+\sqrt{2z}}.$$
Subtracting $\sqrt{2z}$ from both sides, yields to $t  -\sqrt{2z} \le z + \log(1+\sqrt{2z}) -\sqrt{2z} + \frac{t-\sqrt{2z}}{1+\sqrt{2z}}.$
Gathering $t$-dependent terms on the LHS and simplifying, yields
\begin{equation}
t   \le  \sqrt{2z} +\frac{(1+\sqrt{2z})  (z + \log(1+\sqrt{2z}) -\sqrt{2z})}
{\sqrt{2z}} =  \sqrt{2z} +G(z). \label{eq:222}
\end{equation}
Substituting $t$ and $z$  into (\ref{eq:222}), we obtain $\frac{\Lambda-Y}{Y} \le \sqrt{ \frac{2\epsilon}{Y}} + G\left(\frac{2\epsilon}{Y}\right)$ and hence \begin{eqnarray}
\label{L3_5}
\Lambda-Y \le \sqrt{ 2\epsilon Y}+ Y G\left(\frac{2\epsilon}{Y}\right).
\end{eqnarray}
Thus (\ref{L3_5}) gives a bound for $\mathit{Case}$ $\mathit{1}$.

$\mathit{Case}$ $\mathit{2:}$ $\Lambda<Y$.
We start again by rewriting the bound in (\ref{L3_1}) as:
\begin{eqnarray}
\label{eq:tmp:77}
-\log(1- (Y-\Lambda)/Y)-(Y-\Lambda)/Y \le \epsilon/Y.
\end{eqnarray}
Let $t=(Y-\Lambda)/Y$ and $z = \epsilon/Y$ and note that $0 \le t \le 1$. Rewriting inequality (\ref{eq:tmp:77}) in terms of $t$ and $z$, we obtain $-\log(1- t)-t \le z.$
Since $t^2/2 \le -\log(1- t)-t$ for $0 \le t < 1$, we conclude that $t^2/2   \le z ,$
and consequently, $t \le \sqrt{ 2z}$. Substituting $t$ and $z$ back, we obtain $(Y-\Lambda)/Y \le \sqrt{ 2\epsilon /Y}.$
Finally, multiplying both sides by $Y$, we obtain a bound on $Y-\Lambda$: $Y-\Lambda \le \sqrt{ 2\epsilon Y}.$ This gives a bound for $\mathit{Case}$ $\mathit{2}$.

To combine bounds for two cases together, we notice that $|\Lambda-Y| \le \sqrt{ 2\epsilon Y} + Y G\left(\frac{2\epsilon}{Y}\right)$ for $\Lambda>Y$ and $|Y-\Lambda| \le \sqrt{ 2\epsilon Y}$ for $\Lambda<Y$. Therefore, we can use the largest of the two bounds as an upper bound
$|\Lambda-Y| \le \sqrt{ 2\epsilon Y}+  Y G\left(\frac{2\epsilon}{Y}\right).$

\endproof

\begin{lem}
\label{Lemma4}

Let $y,\lambda \in \mathbb{R}^{M\times 1}$ be such that $y(j), \lambda(j) \ge 0$, $\forall j = \overline{1,M}$. Let also, $Y=\|y\|_1$ and $\Lambda=\|\lambda\|_1$. Let $\epsilon >0$ be given, such that $I(y||\lambda) \le \epsilon.$ Then the following inequality holds:
\begin{equation}
\label{lem4bound}
\|y-\lambda\|_1 \le 2\sqrt{2\epsilon Y}+YG\left(\frac{2\epsilon}{Y}\right),
\end{equation}
where auxiliary function $G$ is define as $G(z) = \frac{(1+\sqrt{2z})(z+\log(1+\sqrt{2z})-\sqrt{2z})}{\sqrt{2z}}.$
\end{lem}
\proof
First notice that the norm $\|y-\lambda\|_1$ can be bounded as follows:
$$\|y-\lambda\|_1 = \Big\|Y\Big(\frac{y}{Y}-\frac{\lambda}{Y}\Big)\Big\|_1 = \Big\|Y\Big(\frac{y}{Y}-\frac{\lambda}{Y}+\frac{\lambda}{\Lambda}-\frac{\lambda}{\Lambda}\Big)\Big\|_1 =  \Big\|Y\Big(\Big[\frac{y}{Y}-\frac{\lambda}{\Lambda}\Big]+\Big[\frac{\lambda}{\Lambda}-\frac{\lambda}{Y}\Big]\Big)\Big\|_1$$
$$ \le \Big\|Y\Big(\frac{y}{Y}-\frac{\lambda}{\Lambda}\Big)\Big\|_1+\Big\|Y\Big(\frac{\lambda}{\Lambda}-\frac{\lambda}{Y}\Big)\Big\|_1= Y\Big\|\frac{y}{Y}-\frac{\lambda}{\Lambda}\Big\|_1+|Y-\Lambda|.$$
Now we apply bounds (\ref{l1normalizedbound}) and (\ref{l1YLambda}) obtained in Lemmas (\ref{Lemma2}) and (\ref{Lemma3}), respectively:
$$\|y-\lambda\|_1 \le Y\Big\|\frac{y}{Y}-\frac{\lambda}{\Lambda}\Big\|_1+|Y-\Lambda| \le Y\sqrt{\frac{2\epsilon}{Y}}+\sqrt{2\epsilon Y}+YG\left(\frac{2\epsilon}{Y}\right) = 2\sqrt{2\epsilon Y}+YG\left(\frac{2\epsilon}{Y}\right).$$ Hence, $\|y-\lambda\|_1 \le 2\sqrt{2\epsilon Y}+YG\left(\frac{2\epsilon}{Y}\right).$
\endproof

\subsubsection{ Proof of Theorem \ref{thmML}}
\proof

{\em Exact row sparsity.}

Let matrix $X^{ML}$ be a solution to MLCC-RSM optimization problem (\ref{CCML}). Notice that $\epsilon_{ML}$ is chosen such that $\mathbb{P}\left(\tilde X \in S_{\epsilon_{ML}}\right)\ge 1-p$. Therefore by Proposition \ref{propUpperBound}: $\|X^{ML}\|_{R^0} \le \|\tilde X\|_{R^0}$ with probability at least $1-p$.

Now, since $X^{ML}$ is the solution to MLCC-RSM optimization problem (\ref{CCML}) then $X^{ML}\in S_{\epsilon_{ML}}$. Hence, the Frobenius norm of the difference between the solution matrix $X^{ML}$ and the true row sparse matrix $\tilde X$ can be bounded as follows:
$$ \| X^{ML}- \tilde X \|_F  =  \sqrt{\sum_{i=1}^{N} \| x^{ML}_i- \tilde x_{i} \|_2^2} \stackrel{\circled{1}}{\le}  \frac{\sqrt{ \sum_{i=1}^{N} \|A_i( x^{ML}_i- \tilde x_{i} )\|_2^2} }{1-\delta_{2s}}$$
$$ \stackrel{\circled{2}}{\le}  \frac{\sqrt{ \sum_{i=1}^{N}(\|A_i x^{ML}_i- y_i \|_2 + \|A_i \tilde x_{i} -y_i\|_2)^2}}{1-\delta_{2s}}
 \stackrel{\circled{3}}{\le}  \frac{\sqrt{ \sum_{i=1}^{N}(\|A_i x^{ML}_i- y_i \|_1 + \|A_i \tilde x_{i} -y_i\|_1)^2}}{1-\delta_{2s}},$$
where $\circled{1}$ is due to fact that $A_i$ satisfies the $2s$-restricted isometry property $\forall i = \overline{1,N}$ and that sparsity of vector $x^{ML}_i- \tilde x_{i}$ is at most $2s$, $\forall i=\overline{1,N}$, i.e., $\|x^{ML}_i- \tilde x_{i}\|_0\le 2s$, $\circled{2}$ is due to the triangle inequality, and $\circled{3}$ is due to the fact that $\|z\|_2\le\|z\|_1$, $\forall z$. Now notice that if $\sum_{i=1}^{N}I(y_i||\lambda_i)\le \epsilon_{ML}$ then $I(y_i||\lambda_i)\le \epsilon_{ML}$, $\forall i = \overline{1,N}.$
Therefore, using result (\ref{lem4bound}) of Lemma (\ref{Lemma4}) we bound norms $\|A_i x^{ML}_i- y_i \|_1$ and $\|A_i \tilde x_{i} -y_i\|_1$ as follows:
$$ \|A_i x^{ML}_i- y_i \|_1 \le 2\sqrt{2\epsilon_{ML} \|y_i\|_1}+\|y_i\|_1G\left(\frac{2\epsilon_{ML}}{\|y_i\|_1}\right), \|A_i \tilde x_{i} -y_i\|_1 \le 2\sqrt{2\epsilon_{ML} \|y_i\|_1}+\|y_i\|_1G\left(\frac{2\epsilon_{ML}}{\|y_i\|_1}\right).$$
Therefore, $\| X^{ML}- \tilde X \|_F \le$
$$\frac{\sqrt{ \sum_{i=1}^{N}\Big(\|A_i x^{ML}_i- y_i \|_1 + \|A_i \tilde x_{i} -y_i\|_1\Big)^2}}{1-\delta_{2s}}
\le  \frac{\sqrt{ \sum_{i=1}^{N}\left(4\sqrt{2\epsilon_{ML} \|y_i\|_1}+2\|y_i\|_1G\left(\frac{2\epsilon_{ML}}{\|y_i\|_1}\right)\right)^2}}{1-\delta_{2s}} $$
$$ = \frac{2\sqrt{ \epsilon_{ML}\sum_{i=1}^{N}\left(2\sqrt{2 \|y_i\|_1}+\frac{\|y_i\|_1}{\sqrt{\epsilon_{ML}}}G\left(\frac{2\epsilon_{ML}}{\|y_i\|_1}\right)\right)^2}}{1-\delta_{2s}} < \gamma_{ML}(\tilde X).$$
Hence, it follows that since $X^{ML}\in S_{\epsilon_{ML}}$ then $X^{ML}$ is in the $\gamma_{ML}(\tilde X)$-Frobenius ball neighborhood of $\tilde X$, i.e., $X^{ML}\in \mathrm{B}_{\gamma_{ML}(\tilde X)}^{F}(\tilde X).$ Therefore, $S_{\epsilon_{ML}} \subseteq \mathrm{B}_{\gamma_{ML}(\tilde X)}^{F}(\tilde X).$ Hence, by Proposition \ref{propLowerBound}: $\|X^{ML}\|_{R^0}\ge \|\tilde X\|_{R^0}.$

Combining inequalities $\|X^{ML}\|_{R^0} \le \|\tilde X\|_{R^0}$ and $\|X^{ML}\|_{R^0}\ge \|\tilde X\|_{R^0}$ together, we conclude that $\|X^{ML}\|_{R^0} = \|\tilde X\|_{R^0}$ with probability at least $1-p$.

{\em Exact row sparsity pattern.}

Now, suppose that row sparsity pattern of the solution matrix $X^{ML}$ is different from the row sparsity pattern of $\tilde X$, i.e., $\mathrm{RSupp}(X^{ML}) \ne \mathrm{RSupp}(\tilde X),$ even if row sparsity is the same, $\|X^{ML}\|_{R^0} = \|\tilde X\|_{R^0}.$  Without loss of generality, we can assume that row-support of $X^{ML}$ and $\tilde X$ are different only by one row index:
$\mathrm{RSupp}(X^{ML}) \setminus \mathrm{RSupp}(\tilde X) = \{l_1,l_2\}$, i.e., $l_1$ row of $X^{ML}$ is zero, but $l_1$ row of $\tilde X$ is not zero and $l_2$ row of $X^{ML}$ is not zero, but $l_2$ row of $\tilde X$ is zero.
Now consider Frobenius norm of the difference between the solution matrix $X^{ML}$ and the true row sparse matrix $\tilde X$: $\| X^{ML} - \tilde X \|_F  = \sqrt{\sum_{l=1}^{K} \sum_{i=1}^{N} |X^{ML}(l,i)- \tilde X(l,i) |^2} = \sqrt{\sum_{l=1}^{K} \| ((x^{ML})^l)^{T}- (\tilde x^l)^{T} \|_2^2}.$
From the definition of $\gamma_{ML}(\tilde X)$, it follows that inequality $\| ((x^{ML})^{l_1})^{T}- (\tilde x^{l_1})^{T} \|_2^2 = \| (\tilde x^{l_1})^{T} \|_2^2 \ge (\gamma_{ML}(\tilde X))^2$ holds for $l_1$ term of the sum $\sum_{l=1}^{K} \| ((x^{ML})^l)^{T}- (\tilde x^l)^{T} \|_2^2$. Therefore,
$$\| X^{ML}- \tilde X \|_F =\!\! \sqrt{\sum_{l=1}^{K} \| ((x^{ML})^l)^{T}- (\tilde x^l)^{T} \|_2^2} \ge \sqrt{\| ((x^{ML})^{l_1})^{T}- (\tilde x^{l_1})^{T} \|_2^2} \ge \sqrt{(\gamma_{ML}(\tilde X))^2} = \gamma_{ML}(\tilde X).$$
This contradicts the assumption of the Proposition \ref{propLowerBound} that $\| X^{ML}- \tilde X \|_F<\gamma_{ML}(\tilde X)$ and hence contradicts the assumption that $\|X^{ML}\|_{R^0} = \|\tilde X\|_{R^0}.$ Therefore, $\mathrm{RSupp}(X^{ML}) = \mathrm{RSupp}(\tilde X).$
\endproof

\bibliography{poisson_noise}
\bibliographystyle{ieeetr}

\end{document}